\documentclass{amsart}

\usepackage{amsmath}
\usepackage{amssymb} 
\usepackage{amsthm,amsbsy}
\usepackage{url}            
\usepackage{amsfonts}       
\usepackage{microtype}      
\usepackage{lipsum}
\usepackage{color}
\usepackage[linesnumbered,ruled,vlined]{algorithm2e}
\SetKwInput{KwInput}{Input}                
\SetKwInput{KwOutput}{Output}              
\usepackage{bbm}

\usepackage{graphicx} 
\usepackage{multirow}
\usepackage{float,lscape}
\usepackage{algorithmic}

\title[DTCCA]{Deep Tensor CCA for Multi-view Learning}
\author[H.S. Wong]{Hok Shing Wong}
\author[L. Wang] {Li Wang}
\author[R. Chan]{Raymond Chan}
\author[T. Zeng]{Tieyong Zeng}
\thanks{
		Hok Shing Wong is with the Department
		of Mathematics, The Chinese University of Hong Kong, Hong Kong. E-mail: hswong@math.cuhk.edu.hk. \\
		\indent Li Wang is with the Department
		of Mathematics and Department of Computer Science and Engineering, University of Texas at Arlington, Texas, 76019 USA. E-mail: li.wang@uta.edu. Corresponding Author. \\
		\indent Raymond Chan is with the Department
		of Mathematics, City University of Hong Kong, Hong Kong. E-mail: rchan.sci@cityu.edu.hk.\\
		\indent Tieyong Zeng is with the Department
		of Mathematics, The Chinese University of Hong Kong, Hong Kong. E-mail: zeng@math.cuhk.edu.hk. 
	}

\begin{document}
\maketitle

\begin{abstract}
We present Deep Tensor Canonical Correlation Analysis (DTCCA), a method to learn complex nonlinear transformations of multiple views (more than two) of data such that the resulting representations are linearly correlated in high order. The high-order correlation of given multiple views is modeled by covariance tensor, which is different from most CCA formulations relying solely on the pairwise correlations. Parameters of transformations of each view are jointly learned by maximizing the high-order canonical correlation. To solve the resulting problem, we reformulate it as the best sum of rank-$1$ approximation, which can be efficiently solved by existing tensor decomposition method.
DTCCA is a nonlinear extension of tensor CCA (TCCA) via deep networks. The transformations of DTCCA are parametric functions, which are very different from implicit mapping in the form of kernel function. Comparing with kernel TCCA, DTCCA not only can deal with arbitrary dimensions of the input data, but also does not need to maintain the training data for computing representations of any given data point. Hence, DTCCA as a unified model can efficiently overcome the scalable issue of TCCA for either high-dimensional multi-view data or a large amount of views, and it also naturally extends TCCA for learning nonlinear representation. Extensive experiments on three multi-view data sets demonstrate the effectiveness of the proposed method.
\end{abstract}

\section{Introduction}
Multi-view learning \cite{zhao2017multi,xu2013survey} has been receiving increased attention in many scientific domains since data sets are usually sampled from diverse variables of each object. Due to their heterogeneous properties, these variables can be naturally partitioned into groups. Each group of variables is referred to as a view. Such data sets with multiple views collectively are referred to as multi-view data sets, such as text content of each web page and the anchor text of other web pages linking to this page in web page classification \cite{blum1998combining} and various descriptors used to extract features of each image for image classification \cite{chua2009nus}.

Subspace-based multi-view learning as one of the most representative categories in the multi-view learning paradigm has been extensively studied for high-dimensional multi-view data sets \cite{xu2013survey}. It aims to obtain a latent subspace shared by multiple views based on the assumption that each view of the data is generated from the unknown distribution conditioned on the same latent subspace \cite{bach2005probabilistic}. The ``curse of dimensionality'' can be effectively alleviated by learning a latent subspace with the dimensionality less than any of the input views. Canonical correlation analysis (CCA), originally designed for measuring the linear correlation between two sets of variables \cite{harold1936relations}, was formally introduced as a multi-view dimensionality reduction method in \cite{foster2008multi} for its ability of reducing the labeled instance complexity under certain weak assumptions. Another appealing property of CCA is that the learned subspace will not contain the noise in the uncorrelated dimensions if there is noise in either view that is uncorrelated to the other view \cite{andrew2013deep}.
In the last decade, CCA has received a renewed interest in the machine learning community \cite{andrew2013deep,hardoon2004canonical,nielsen2002multiset}, and in many scientific fields its usefulness and those of its variants have already been well demonstrated \cite{uurtio2017tutorial}.

We in this paper are particularly interested in the multi-view data sets with more than two views and the inherent nonlinear property between the latent subspace and the input views. Two representative nonlinear representation techniques have been applied to CCA for two views: kernel trick \cite{hardoon2004canonical} and deep learning \cite{andrew2013deep}. Kernel CCA (KCCA) \cite{hardoon2004canonical} extends CCA for finding maximally correlated nonlinear projections in reproducing kernel Hilbert space (RKHS) \cite{scholkopf2002learning}. The nonlinearity of KCCA is represented by kernel function, so this representation is limited by the fixed kernel. Moreover, the kernel trick increases the time complexity for learning the projections and computing the representation of new data points since it scales poorly with the size of the training data. To overcome the above drawbacks, deep CCA (DCCA) \cite{andrew2013deep} was proposed by simultaneously learning two deep nonlinear mappings of two views that are maximally correlated. Since the deep networks are parametric and not limited to RKHS, it does not face the above drawbacks of kernel trick, and they have showed the empirical success on various tasks \cite{hinton2006reducing}.  Some variants of DCCA have been studied including deep canonically correlated autoencoders \cite{wang2015deep} by simultaneously maximizing canonical correlation and minimizing the reconstruction errors of the autoencoders, deep variational CCA \cite{wang2016deep} extended from variational autoencoders \cite{kingma2013auto} based on the probabilistic CCA model \cite{bach2005probabilistic}, and deep discriminative CCA \cite{elmadany2016multiview} by considering one labeled data as one view in the setting of supervised learning.

The aforementioned CCA variants are mainly designed for data sets of two views. Various learning criteria have been proposed to extend CCA for more than two views. In work \cite{nielsen2002multiset, kettenring1971canonical}, five multiset correlation formulations and four sets of constraints are discussed for multiset CCAs. Among them, CCA with the sum of pairwise correlations (SUMCOR) criterion enjoys a nice analytic solution by generalized eigen-decomposition, which was reformulated as the least square problem in order to develop an adaptive learning algorithm \cite{via2007learning}. Generalized CCA (GCCA) \cite{horst1961generalized} takes a different perspective by learning a common representation and imposing orthogonality on the common representation. We will build the connection between GCCA and CCA with SUMCOR and show in Section \ref{sec:mcca-extension} that GCCA does not really maximize the canonical correlation so it does not reduce to CCA for two views. For more than two views, multiset CCA and GCCA can only capture the pairwise correlations. To generalize CCA for handling more than two views, tensor CCA (TCCA) \cite{luo2015tensor} was proposed by maximizing the high-order correlation represented by the covariance tensor \cite{tao2007general} over the data sets from all views, so it is a natural way to extend CCA for arbitrary number of views. Nonlinear extensions of these multi-view CCAs have also been explored. Kernel TCCA extends CCA based on kernel trick, so it encounters the same drawbacks  as KCCA. Deep multiset CCA extends multiset CCA via deep networks, but it only can deal with very special case of multi-view data sets where views have to be sampled from the same input space. Deep GCCA (DGCCA) extends GCCA via deep networks, but it does not reduce to DCCA for two views.

In this paper, we propose deep TCCA (DTCCA) by extending TCCA for learning nonlinear projections via deep networks. DTCCA not only inherits the high-order canonical correlation of multiple views but also overcomes the drawbacks brought by kernel TCCA. 
The main contributions of this paper are summarized as follows:
\begin{itemize}
\item We build the connections among three representative categories of existing CCAs for more than two views and their nonlinear generalizations. Based on the connections, the pros and cons of these methods are discussed in detail.

\item We further propose DTCCA model which can simultaneously learn the nonlinear projections and TCCA via deep networks. Comparing to kernel TCCA, DTCCA can effectively overcome the drawbacks caused by kernel function and make TCCA practical for large-scale and high-dimensional multi-view data sets. To the best of our knowledge, there is no prior work on the nonlinear generalization of TCCA via deep networks.

\item Extensive experiments are performed on three multi-view data sets by comparing with various representative baselines in terms of various settings including varied views, the dimensions of latent subspace, and the ratios of training data. Moreover, the impact on the number of layers of networks are also investigated. Experimental results show that DTCCA significantly outperform TCCA and other methods especially on small amount of training data, and it shows consistent better results over data sets with more than two views and varied latent dimensions.
\end{itemize}

In the following of this paper, we first review CCA and their multi-view extensions including nonlinear generalization. In Section \ref{sec:dtcca}, the proposed model is presented, and the optimization problem is reformulated as tensor decomposition. Extensive experiments are conducted in Section \ref{sec:experiments}. Finally, we draw our conclusions in Section \ref{sec:conclusions}.

\section{CCA and Its Multi-view Extensions} \label{sec:mcca-extension}      
CCA was originally proposed to find a pair of bases for two set of variables so that their corresponding projected variables onto these bases are maximally correlated \cite{harold1936relations}. Its generalization to multiple data sets (more than two views) has been widely studied due to the emerging of multi-view data sets in various real world applications. In this section, we will review three representative multi-view CCA methods and their nonlinear extensions according to their inherent criteria.

Denote by $\{ (\mathbf{x}_1^i,\ldots,\mathbf{x}_k^i)\}_{i=1}^n$ the data sets of $k$ views with $n$ data points, where the $i$th data of the $r$th view is $\mathbf{x}^i_r \in \mathbb{R}^{d_r}$ and $d_r$ is the dimension of the $r$th view. Let $X_r = [\mathbf{x}_r^1,\ldots,\mathbf{x}_r^n] \in \mathbb{R}^{d_r \times n}$ be the matrix representation of the $r$th view data set and $P_r = [\mathbf{p}_r^1,\ldots,\mathbf{p}_r^m] \in \mathbb{R}^{d_r \times m}$ be the projection matrix for the $r$th view consisting of $m$ bases in $d_r$-dimensional space. Denote $P=[P_1;\ldots;P_k] \in \mathbb{R}^{d \times m}$ with $d=\sum_{r=1}^k d_r$.
Without loss of the generality, we assume that data of each view is centered, that is, $ X_r \mathbf{1}_n = \mathbf{0}_{d_r}, \forall r=1,\ldots,k$, where $\mathbf{1}_n$ is an $n$-dimensional column vector of all $1$s and $\mathbf{0}_{d_r}$ is a $d_r$-dimensional column vector of all $0$s. The cross-view covariance between view $r$ and view $s$ is defined  as $C_{r,s} = X_r X_s^T\in \mathbb{R}^{d_r \times d_s}$ and the intra-view covariance of view $r$ is defined as $C_{r,r} =X_r X_r^T\in \mathbb{R}^{d_r \times d_r}$. $\|A\|_F$ is the Frobenius norm of matrix $A$.

\subsection{Multiset CCA}
Multiset CCA (MCCA) \cite{nielsen2002multiset, kettenring1971canonical} has been studied for analyzing linear relations between more than two views, and various formulations have been explored. The straightforward extension of CCA to multiset CCA is to maximize the sum of the pairwise correlations (CCA-SUMCOR):
\begin{align}
\max_{\{P_r\}_{r=1}^k} & \sum_{r=1}^k \sum_{s=1}^k \textrm{tr} ( P_r^T C_{r,s}  P_s)
~\textrm{s.t.}  \sum_{r=1}^k P_r^T C_{r,r}  P_r = I_m,  \label{op:sumcor}
\end{align}
where the orthogonal constraint over the projected data is added to prevent trivial solution. For $k=2$, problem (\ref{op:sumcor}) is reduced to the conventional CCA problem. The Langrange multiplier technique can be used to solve the above constrained maximization problem. With the multiplier diagonal matrix $\Lambda$, we can formulate the Lagrangian function $L(\{P_r\}, \Lambda)$ as
\begin{align}
\sum_{r=1}^k \sum_{s=1}^k \textrm{tr} ( P_r^T C_{r,s}  P_s) -  \textrm{tr}\left(\Lambda (\sum_{r=1}^k P_r^T C_{r,r}  P_r - I_m)\right).
\end{align}
The optimality condition is
\begin{align}
\sum_{s=1}^k C_{r,s} P_s =  C_{r,r} P_r \Lambda, \forall r=1,\ldots, s,
\end{align}
which is equivalent to the following matrix representation
\begin{align}
\begin{bmatrix}
C_{1,1} & C_{1,2} & \ldots & C_{1,k}\\
C_{2,1} & C_{2,2} & \ldots & C_{2,k}\\
\vdots & \vdots & \ddots &  \ldots\\
C_{k,1} & C_{k,2} & \ldots & C_{k,k}\\
\end{bmatrix} \!\! P\!\!
= \!\!
\begin{bmatrix}
C_{1,1} & 0 & \ldots & 0\\
0 & C_{2,2} & \ldots & 0\\
\vdots & \vdots & \ddots &  \ldots\\
0 & 0 & \ldots & C_{k,k}\\
\end{bmatrix} \! P\! \Lambda \label{eq:sumcor-geig}.
\end{align}
It is well-known that the optimal projections $\{P_r\}_{r=1}^k$ can be obtained by solving the generalized eigenvalue decomposition problem (\ref{eq:sumcor-geig}). Moreover, problem (\ref{op:sumcor}) can be equivalently rewritten as
\begin{align}
\!\!\min_{\{P_{r}\}}  \sum_{r=1}^k \sum_{s=1}^k\! \| P_r^T X_r - P_s^T X_s \|_F^2 ~\textrm{s.t.}  \sum_{r=1}^k P_r^T C_{r,r}  P_r = I_m, \label{op:LSCCA}
\end{align}
since $\sum_{r=1}^k \sum_{s=1}^k \| P_r^T X_r\|_F^2 = k \textrm{tr}(\sum_{r=1}^k P_r^T C_{r,r}  P_r) = km$ is a constant. The pairwise least square formulation (\ref{op:LSCCA}) (LSCCA) was proposed to develop an adaptive learning algorithm based on the recursive least squares \cite{via2007learning}. Another reformulation can be achieved by introducing the average representation of the $k$ views denoted by
\begin{align}
M = \frac{1}{k} \sum_{r=1}^k P_r^T X_r, \label{eq:mean}
\end{align}
and we have the following equalities:
\begin{align}
& \sum_{r=1}^k \sum_{s=1}^k \| P_r^T X_r - P_s^T X_s \|_F^2 \nonumber\\
=& \sum_{r=1}^k \sum_{s=1}^k \| (P_r^T X_r - M )- (P_s^T X_s-M) \|_F^2 \nonumber\\
=& 2k \sum_{r=1}^k \| P_r^T X_r - M \|_F^2. \label{eq:center}
\end{align}
Accordingly, the reformulated problem of (\ref{op:sumcor}) based on (\ref{eq:mean}) and (\ref{eq:center}) is written as
\begin{align}
\min_{\{P_r\}_{r=1}^k, M} & \sum_{r=1}^k \| P_r^T X_r -  M \|_F^2 ~\textrm{s.t.}  \sum_{r=1}^k P_r^T C_{r,r}  P_r = I_m. \label{op:maxvar}
\end{align}
The equivalence between LSCCA and CCA-MAXVAR \cite{kettenring1971canonical} has been proved in  \cite{via2007learning}, so (\ref{op:maxvar}) is also equivalent to LSCCA and CCA-MAXVAR.
Problem (\ref{op:sumcor}) with alternating constraints $P_r^T C_{r,r} P_r = I_m, \forall r$ was also explored in \cite{nielsen2002multiset}, but it loses the nice analytic solution (\ref{eq:sumcor-geig}) that  (\ref{op:sumcor}) has. Moreover, the supervised extension of MCCA has also been explored by incorporating label data via linear discriminant analysis \cite{sun2015multiview,cao2017generalized}.

Deep CCA (DCCA) \cite{andrew2013deep} is proposed to learn two nonlinear transformations $f_1$ and $f_2$ by simultaneously maximizing the correlation between two views:
\begin{align}
\max_{f_1, f_2, P_1, P_2} \frac{\textrm{tr}(P_1^T \hat{C}_{1,2} P_2)}{\sqrt{ \textrm{tr}(P_1^T \hat{C}_{1,1} P_1) \textrm{tr}(P_2^T \hat{C}_{2,2} P_2) }}, \label{op:CCA}
\end{align}
where $f_1$ and $f_2$ can be multiple stacked layers of nonlinear transformations with output dimension as $m$, $\hat{C}_{1,2} = \hat{f}_1(X_1) \hat{f}_2(X_2)^T$, $\hat{C}_{r,r} = \hat{f}_r(X_r) \hat{f}_r(X_r)^T, \forall r=1, 2$, and $\hat{f}_r(X_r) = f_r(X_r) H$ is a centered transformed matrix with centering matrix $H = I_n - \frac{1}{n} \mathbf{1}_n \mathbf{1}_n^T \in \mathbb{R}^{n \times n}$. Let $\hat{P}_1 = \hat{C}_{1,1}^{1/2} P_1$ and $\hat{P}_2 = \hat{C}_{2,2}^{1/2} P_2$. Problem (\ref{op:CCA}) is equivalent to the following maximization problem
\begin{align}
\max_{f_1, f_2, \hat{P}_1, \hat{P}_2} \textrm{tr}(\hat{P}_1^T T\hat{P}_2) ~\textrm{s.t.}~ \hat{P}_1^T \hat{P}_1 = \hat{P}_2^T \hat{P}_2 = I_m, \label{op:CCA-svd}
\end{align}
where $T=\hat{C}_{1,1}^{-1/2} \hat{C}_{1,2} \hat{C}_{2,2}^{-1/2} $. For fixed $f_1$ and $f_2$, problem (\ref{op:CCA-svd}) can be solved optimally by singular value decomposition (SVD). Let $U_m$ and $V_m$ be the matrices of the top $m$ left- and right- singular vectors of $T$. We have the optimal solution $P_1 = \hat{C}_{1,1}^{-1/2} U_m$ and $P_2 = \hat{C}_{2,2}^{-1/2} V_m$. And the optimal objective becomes $\textrm{tr}(T^T T)^{1/2}$, which is a function of $f_1$ and $f_2$. The work \cite{wang2015deep} further explored the autoencoder to regulate DCCA. Unfortunately, the special reformulation (\ref{op:CCA-svd}) for CCA is not applicable for MCCA since $n^2$ pairs of projections are coupled. dMCCA \cite{somandepalli2019multimodal} extends MCCA for nonlinear transformation via deep networks, but it only can deal with very special case that all view data sets are sampled from the same input space, that is $d_r = d_s, \forall s\not=r$, and $P_r=P_s, \forall r\not=s$. These strong assumptions prevent dMCCA from being used for general multi-view data sets.

\subsection{Generalized CCA} 
Generalized CCA (GCCA) \cite{horst1961generalized} finds $\{ P_r \}_{r=1}^k$ by constructing a common representation $G \in \mathbb{R}^{m \times n}$ so that the sum of the squared losses between each view and $G$ is minimized. GCCA is formulated as the optimization problem
\begin{align}
\min_{G, \{P_r\}_{r=1}^k} &~\sum_{r=1}^k \| G - P_r^T X_r \|_F^2 ~\textrm{s.t.} ~ G G^T = I_m, \label{op:GCCA}
\end{align}
where the orthogonal constraint over $G$ is added so as to prevent the trivial solution. It is worth noting that (\ref{op:GCCA}) resembles (\ref{op:maxvar}) in terms of the objective function, but the constraints are very different.
Problem (\ref{op:GCCA}) can be transformed to an eigenvalue decomposition problem.  First, given a matrix $G$, problem (\ref{op:GCCA}) with respect to $P_r$ can be solved independently, and it is a convex quadratic programming so it can be solved globally by the first order optimality condition, that is
\begin{align}
-2 X_r (G - P_r^T X_r)^T   = 0 \Rightarrow P_r = (X_r X_r^T)^{-1} X_r G^T. \label{eq:GCCA-P}
\end{align}
By substituting (\ref{eq:GCCA-P}) back to (\ref{op:GCCA}), we then reformulate (\ref{op:GCCA}) as
\begin{align}
\max_{G}~ \textrm{tr} ( G Q G^T ) ~\textrm{s.t.}~ G G^T = I_m, \label{op:GCCA-eig}
\end{align}
where $Q = \sum_{r=1}^k X_r^T  (X_r X_r^T)^{-1} X_r$. Hence, the optimal solution $G$ consists of the eigenvectors  corresponding to the top $m$ eigenvalues of $Q$. Once $G$ is obtained, $\{P_r\}_{r=1}^k$ can be recovered by (\ref{eq:GCCA-P}). Some other extensions of GCCA have also been explored, such as the $\ell_{2,1}$-norm regularized GCCA model is proposed to facilitate the interpretability of the learning representation \cite{xu2019canonical}.

Deep GCCA (DGCCA) \cite{Benton2019} extends (\ref{op:GCCA}) for nonlinear multi-view learning, where the input data of each view is replaced by the transformed data via some nonlinear function $\{f_r\}_{r=1}^k$, e.g., the multi-layer perception network. This can effectively resolve the drawback of GCCA for only learning linear projections. The optimization problem is formulated as
\begin{align}
\min_{G, \{P_r\}_{r=1}^k, \{f_r\}_{r=1}^k} &~\sum_{r=1}^k \| G - P_r^T f_r(X_r) \|_F^2 \label{op:DGCCA}\\
\textrm{s.t.} &~ G G^T = I_m. \nonumber
\end{align}
The same reformulated problem as GCCA is obtained 
\begin{align}
\max_{G, \{f_r\}_{r=1}^k}~ \textrm{tr} ( G Q_f G^T ) ~\textrm{s.t.}~ G G^T = I_m, \label{op:DGCCA-eig}
\end{align}
where $Q_f = \sum_{r=1}^k f_r(X_r)^T  (f_r(X_r) f_r(X_r)^T)^{-1} f_r(X_r)$. The gradient with respect to $f_r$ can be calculated as $\partial_{f_r} = 2 P_r G - 2 P_r P_r^T f_r(X_r)$.

\subsection{Tensor CCA} \label{sec:tcca}
Tensor CCA (TCCA) \cite{luo2015tensor} is proposed for multi-view learning by exploiting high-order tensor correlation among multiple views. Let $\mathbf{z}_r^l = [z_r^l(1),\ldots,z_r^l(n)]^T= X_r^T \mathbf{p}_r^l \in \mathbb{R}^n$ be the canonical variable of the $r$th view projected onto  the $l$th base. The high-order canonical correlation over $k$ views is defined as
\begin{align}
\rho &= \sum_{l=1}^m \rho_l  \label{eq:tcc}\\
\rho_l &= \sum_{i=1}^n \prod_{r=1}^k  z_r^l(i), \forall l=1,\ldots,m, \label{eq:high-order-correlation}
\end{align}
with the constraints used in \cite{kettenring1971canonical}
\begin{align}
P_r^T C_{r,r}  P_r = I_m, \forall r=1,\ldots,m. \label{eq:con}
\end{align}
In the case of $k=2$, the high-order correlation is reduced to the canonical correlation, which can be verified by the following derivations:
\begin{align*}
\textrm{tr} ( P_1^T X_1 X_2^T  P_2) =& \sum_{l=1}^m (\mathbf{z}_1^l)^T \mathbf{z}_2^l \\
=& \sum_{l=1}^m \sum_{i=1}^n z_1^l(i) z_2^l(i) = \rho.
\end{align*}
As a result, maximizing (\ref{eq:tcc}) is equivalent to CCA-MAXVAR in the case of $k=2$ and constraints (\ref{eq:con}). For $k>2$, the high-order correlation will be captured by (\ref{eq:tcc}). For multi-view learning, this makes TCCA different from others based on pairwise correlations. As a result, maximizing (\ref{eq:tcc}) is transformed to the best sum of rank-1 approximation, e.g., the best rank-$m$ CANDECOMP/PARAFAC decomposition \cite{carroll1970analysis}. 
The well-known alternating least squares (ALS) algorithm \cite{kroonenberg1980principal,comon2009tensor} is used. The nonlinear extension of TCCA via kernel trick is also explored in \cite{luo2015tensor}.  

\subsection{Discussions}
We are now ready to compare the above three representative multi-view extensions of CCA from two perspectives: learning criterion and the nonlinear extension.

As the learning criterion, their correlation definitions are different.
Both MCCA and TCCA generalize CCA since they reduce to exact CCA for two views. However, GCCA does not possess this property. By comparing (\ref{op:maxvar}) with (\ref{op:GCCA}), it is easy to see that GCCA enforces orthogonality on the common representations, while MCCA takes the mean of all view representations (\ref{eq:mean}) as the common representation. This implies that GCCA is suitable for visualizing multi-view data in the orthogonal coordinate space, while MCCA is good to maximize the pairwise correlation of any two views by assuming the common representation variable as the mean of projected data of all views. In contrast, TCCA is very different from MCCA for $k > 2$ and GCCA since the high-order correlations among views can be captured by TCCA, but MCCA can be only able to model the pairwise correlation and GCCA only captures the linearly transformed intra-view correlation according to (\ref{eq:GCCA-P}) and (\ref{op:GCCA-eig}).

Two techniques are popularly used for learning nonlinear projections of multi-view data: kernel trick and deep networks. Kernel CCA (KCCA) \cite{hardoon2004canonical} models the nonlinear transformation via kernel functions. Kernel TCCA extends TCCA via kernel trick. However, kernel trick faces two crucial issues: restricted representation power of a fixed kernel function and the high-computational complexity for large-scale data, even though kernel learning \cite{gonen2011multiple} and kernel approximation \cite{arora2012kernel} techniques have been studied. Fortunately, DCCA \cite{andrew2013deep} effectively alleviates the two issues for nonlinear CCA by learning deep networks as the nonlinear transformation functions. For multi-view CCAs, the nonlinear representation learning is still limited. For example, dMCCA is only applicable for special data sets and DGCCA does not align well with correlation maximization. In this paper, we propose to extend TCCA for nonlinear projections using deep networks, which not only inherits the advantages of TCCA but also makes nonlinear representation practical by deep networks. To the best of our knowledge, there is no prior work on extending TCCA to nonlinear transformation using deep networks.

\begin{figure}
\centering
\includegraphics[width=0.85\textwidth]{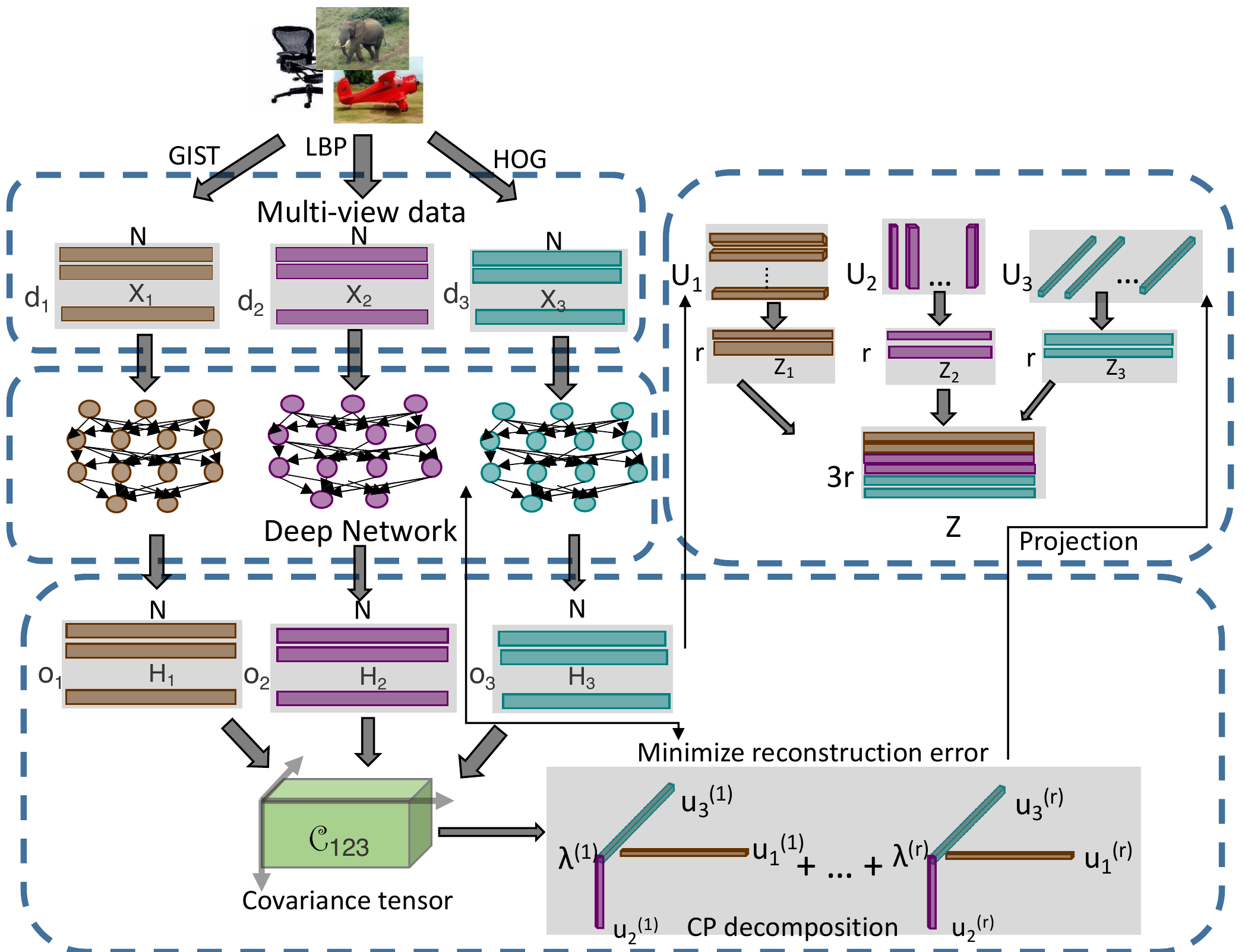}
\caption{The workflow of the proposed DTCCA model by maximizing three-order covariance defined by covariance tensor via independent deep networks with illustrated example consisting of three views from descriptors GIST, LBP and HOG.} \label{fig:workflow}
\end{figure}

\section{Deep Tensor CCA} \label{sec:dtcca}
In this paper, we propose Deep Tensor CCA (DTCCA), which computes the representations of multiple views by passing them through multi-layer perception networks with layers of nonlinear transformations, and the networks are tuned automatically by maximizing the high-order canonical correlation (\ref{eq:tcc}). Fig. \ref{fig:workflow} illustrates the workflow of DTCCA with a data set consisting of three views.

Without loss of the generality, we assume that the $i$th intermediate layer in the network for the $r$th view has $c_r^i$ units, and the output layer has $m$ units. The output of the first layer for the input data $\mathbf{x}_r$ from the $r$th view is $h_r^1=\sigma(W^1_r \mathbf{x}_r +b^1_r) \in \mathbb{R}^{c_r^1}$, where $W^1_r \in \mathbb{R}^{c_r^1 \times d_r}$ is the weight matrix, $b^1_r \in \mathbb{R}^{c_r^1}$ is the vector of biases, and $\sigma: \mathbb{R} \rightarrow \mathbb{R}$ is a nonlinear activation function.
The output $h_r^1$ can then be used as the input to the next layer for computing the output $h_r^2 = \sigma(W^2_r h_r^1 +b^2_r) \in \mathbb{R}^{c_r^2}$, and this is recursively constructed $v$ times until the final output $f_r(\mathbf{x}_r) = \sigma(W^v_r h_r^{v-1} +b^v_r) \in \mathbb{R}^{m}$ is reached. The same construction process can be used for each of the $k$ views. As a result, we have a set of nonlinear functions $\{f_r\}_{r=1}^k$ with the number of intermediate layers $v$ and their associated parameters $\{ W^i_r, b^i_r \}, \forall r=1,\ldots,k, i=1,\ldots,v$. To simplify the annotation, we assume $f_r$ implicitly associates to its network parameters.

With the above defined nonlinear transformation $\{f_r\}_{r=1}^m$, DTCCA aims to maximize the high-order canonical correlation by solving the following optimization problem
\begin{align}
\max_{\{f_r\}_{r=1}^k, \{P_r\}_{r=1}^k} &~ \sum_{l=1}^m \sum_{i=1}^n \prod_{r=1}^k  z_r^l(i) \label{op:dcca} \\
\textrm{s.t.} &~ \mathbf{z}_r^l = \hat{f}_r(X_r)^T \mathbf{p}_r^l, \forall r=1,\ldots,k, l=1,\ldots, m, \nonumber\\
&~ P_r^T \hat{f}_r(X_r) \hat{f}_r(X_r)^T  P_r = I_m, \forall r=1,\ldots,k, \nonumber\\
&~ \hat{f}_r(X_r) = f_r(X_r) H,  \forall r=1,\ldots,k,\nonumber
\end{align}
where $\hat{f}_r(X_r) \in \mathbb{R}^{m \times n}$ is the centered matrix of $f_r(X_r)$. In order to jointly optimize the network and TCCA, we will first transform (\ref{op:dcca}) to the best rank-$m$ tensor decomposition problem. 
Define the covariance tensor of the network output data $\{\hat{f}_r(X_r) \}_{r=1}^k$ as a $k$-order tensor of size $d_1 \times \ldots \times d_k$ 
\begin{align}
\mathcal{C} =\sum_{i=1}^n \hat{f}_1(\mathbf{x}_1^i) \circ \hat{f}_2(\mathbf{x}_2^i) \circ \ldots \circ \hat{f}_k(\mathbf{x}_k^i) 
\end{align}
where $\circ$ is the outer product of vectors.
Let $U \in \mathbb{R}^{p \times d_r}$ be a matrix. The $r$-mode product of tensor $\mathcal{C}$ and $U$ is defined as a tensor $\mathcal{A} = \mathcal{C}\times_r U $ of size $d_1 \times \ldots d_{r-1} \times p \times d_{r+1}\times \ldots \times d_k$ with element
\begin{align*}
\mathcal{A} (i_1,\ldots i_{r-1}, j_r, i_r, \ldots, i_k) = \sum_{i_r=1}^{d_r} \mathcal{C}(i_1,\ldots,i_k) U(j_r, i_r).
\end{align*}
The high-order canonical correlation (\ref{eq:high-order-correlation}) can be rewritten as 
\begin{align}
\sum_{i=1}^n \prod_{r=1}^k  z_r^l(i) = \mathcal{C} \times_1 (\mathbf{p}_1^l)^{\mathrm{T}} \times_2 (\mathbf{p}_2^l)^{\mathrm{T}} \ldots \times_k (\mathbf{p}_k^l)^{\mathrm{T}}.
\end{align}
Similar to DCCA, the orthogonal constraints in (\ref{op:dcca}) can be rewritten as, $\forall r= 1,\ldots, k$
\begin{align}
(\mathbf{p}_r^l )^{\textrm{T}} \hat{C}_{r,r} \mathbf{p}_r^{l'}   = \left\{
\begin{array}{ll}
1, &  l=l' \\
0, & \textrm{otherwise.}
\end{array}
\right.
\end{align}
where  $\hat{C}_{r,r} = \hat{f}_r(X_r) \hat{f}_r(X_r)^T, \forall r=1, \ldots, k.$
Let $\mathbf{u}_r^l = \hat{C}_{r,r}^{1/2} \mathbf{p}_r^l$ and $U_l = [\mathbf{u}_r^1,\ldots, \mathbf{u}_r^m] \in \mathbb{R}^{m \times m}$. We can further reformulate (\ref{op:dcca}) as
\begin{align}
\max_{\{f_r\}_{r=1}^k, \{U_r\}_{r=1}^k} &\! \sum_{l=1}^m \! \mathcal{M} \times_1 (\mathbf{u}_1^l)^{\mathrm{T}} \times_2 (\mathbf{u}_2^l)^{\mathrm{T}} \ldots \times_k (\mathbf{u}_k^l)^{\mathrm{T}}\!\label{op:rank-m}\\
\textrm{s.t.} &~ \mathcal{M} =  \mathcal{C} \times_1 C_{1,1}^{-1/2} \times_2 C_{2,2}^{-1/2} \ldots \times_k C_{k,k}^{-1/2}\nonumber\\
&~
(\mathbf{u}_r^l )^{\textrm{T}} \mathbf{u}_r^{l'}   = \left\{
\begin{array}{ll}
1, &  l=l' \\
0, & \textrm{otherwise.}
\end{array}
\right. \nonumber
\\
&~ \hat{f}_r(X_r) = f_r(X_r) H,  \forall r=1,\ldots,k. \nonumber
\end{align}
Problem (\ref{op:rank-m}) consists of the best sum of rank-1 approximation, e.g., the best rank-$m$ CANDECOMP/PARAFAC decomposition \cite{carroll1970analysis}. This is given by
\begin{align}
\hat{\mathcal{M}} = \sum_{l=1}^m \lambda_l   \mathbf{u}_1^l \circ \mathbf{u}_2^l \ldots \circ \mathbf{u}_k^l.
\end{align}
According to \cite{de2000multilinear}, the sum of rank-1 decomposition and orthogonality constraints of high-order tensor cannot be satisfied simultaneously. Although $U_r$ is not enforced to be orthogonal, $(\mathbf{u}_r^l )^{\textrm{T}} \mathbf{u}_r^{l} =1$ holds. Hence, we resort to solving an approximation problem given by
\begin{align}
\min_{\{f_r\}_{r=1}^k, \{U_r\}_{r=1}^k} &\! \| \mathcal{M} - \hat{\mathcal{M}}\|_F^2 \label{op:als}\\
\textrm{s.t.} &~ \mathcal{M} =  \mathcal{C} \times_1 C_{1,1}^{-1/2} \times_2 C_{2,2}^{-1/2} \ldots \times_k C_{k,k}^{-1/2}\nonumber\\
&~
(\mathbf{u}_r^l )^{\textrm{T}} \mathbf{u}_r^{l}   = 1, \forall r=1,\ldots,k, l=1,\ldots,m, \nonumber
\\
&~ \hat{f}_r(X_r) = f_r(X_r) H,  \forall r=1,\ldots,k, \nonumber
\end{align}
where $\|\mathcal{M}\|_F$ is the Frobenius norm of the tensor $\mathcal{M}$. Given an $\mathcal{M}$, problem (\ref{op:als}) with respect to $\{U_r\}_{r=1}^k$can be solved by the ALS algorithm \cite{kroonenberg1980principal,comon2009tensor}. The parameters of networks are then updated by minimizing the square loss. It is worth noting that for $k=2$, TCCA is equivalent to CCA as shown in Section \ref{sec:tcca}, so DTCCA with $k=2$ is reduced to DCCA. Algorithm \ref{algo:dtcca} for $k=2$ provides an alternative approach for solving DCCA since ALS algorithm obtains an approximate solution, while DCCA takes singular value decomposition during the network learning. However, DCCA approach does not work for $k > 2$.

Once $\{U_r\}_{r=1}^k$ and $\{f_r\}_{r=1}^k$ are obtained after training, we can recover the canonical variables for any given test data $\mathbf{x}_r$ from the $r$th view by
\begin{align}
\mathbf{z}_r = \hat{f}_r(\mathbf{x}_r)^T C_{rr}^{-1/2} U_r, \forall r=1,\ldots,k.
\end{align}

The learning algorithm for DTCCA is presented in Algorithm \ref{algo:dtcca}.
During the training process, we take the full-batch optimization approach, as suggested in \cite{andrew2013deep} for training DCCA. We implement Algorithm \ref{algo:dtcca} in \textit{Pytorch} \cite{NEURIPS2019_9015} together with package \textit{TensorLy} \cite{tensorly}  for tensor operation and decomposition. The Adam optimizer is used with the learning rate set to be $10^{-3}$, and others are set to be default values.

\begin{algorithm}[t]
\caption{Deep tensor CCA (DTCCA)} \label{algo:dtcca}
\begin{algorithmic}[1]
	\STATE \textbf{Input}: data sets of $k$ views: $\{X_r \in \mathbb{R}^{d_r \times n}\}_{r=1}^k$
	\STATE Initialize the networks
	\FOR{$i=1$ to $epoch$}
	\STATE compute $\{f_r(X_r)\}_{r=1}^k$  and then $\{C_{r,r}\}_{r=1}^k$
	\STATE construct tensor $\mathcal{M}$
	\STATE solve rank-$m$ tensor decomposition using ALS algorithm
	\STATE form $\hat{\mathcal{M}}$ based on the solutions $\{\lambda_l\}_{l=1}^m$ and $\{U_r\}_{r=1}^k$
	\STATE update networks by minimizing loss $\| \mathcal{M} - \hat{\mathcal{M}}\|_F^2$
	\ENDFOR
	\STATE $P_r = C_{rr}^{-1/2} U_r, \forall r=1,\ldots, k$
	\STATE \textbf{Output}: $\{f_r\}_{r=1}^k$ and $\{P_r\}_{r=1}^k$.
\end{algorithmic}
\end{algorithm}

\section{Experiments} \label{sec:experiments}
We conduct experiments on three data sets to demonstrate that DTCCA learns nonlinear transformations that not only outperforms TCCA but also shows competitive or better results comparing with other representative models.  Specifically, we compare the following methods in our experiments:
\begin{itemize}
\item TCCA$_p$ \cite{luo2015tensor}. The $m$-order tensor can be high memory intensive for  $m \geq 3$ and moderate-dimensional data sets. To make it applicable for all data sets used in the experiments, we first apply PCA on the input data of each view and reduce their dimensions up to $20$ to make sure TCCA is feasible. As noted in \cite{luo2015tensor} that the kernel version of TCCA does not work for moderate size of data sets with multiple views due to the high memory requirement and computational complexity, so we will not include it in the experiments.

\item LSCCA \cite{via2007learning} and LSCCA+PCA (LSCCA$_p$). As the representative MCCA method, LSCCA  is evaluated on the original input data. In addition, we also preprocess the input data using PCA by preserving $95\%$ energy, which is analogous to the preprocess of TCCA and representation learning of deep learning methods. dMCCA \cite{somandepalli2019multimodal} is not included due to its incapability to the data sets used in the experiments.

\item GCCA \cite{horst1961generalized}, GCCA+PCA (GCCA$_p$), and DGCCA \cite{Benton2019}. Even though GCCA and its variants specifically model the common representation, the learning performance is still evaluated based on the learned projections. The preprocessing using PCA is also applied to GCCA similar to LSCCA$_p$. DGCCA is implemented using the same multi-layer perceptron networks as the one used in the proposed DTCCA method for fair comparisons.

\item DTCCA. The proposed method is implemented as the nonlinear extension of TCCA using multi-layer perceptron networks for each view as shown in Algorithm \ref{algo:dtcca}. DTCCA can handle high-dimensional data and  large-scale data, so the preprocessing step using PCA is not applied. We refer to a DTCCA model with an output size of $m$ and $v$ layers (including output) as DTCCA-$m$-$v$.
\end{itemize}

\begin{table}
\caption{Accuracies of $7$ compared methods on Caltech101 with $6$ views for all $10$ folds.} \label{tab:splits}
\centering
\begin{tabular}{@{}l@{}|c@{\hspace{0.9em}}c@{\hspace{0.9em}}c@{\hspace{0.9em}}c@{\hspace{0.9em}}c@{\hspace{0.9em}}c@{\hspace{0.9em}}c@{}}
	\hline
	&GCCA&LSCCA&TCCA$_p$&GCCA$_p$&LSCCA$_p$&DGCCA&DTCCA\\
	\hline\hline
	Fold 1  & 33.31 & 85.86 & 93.08 & 89.10 & 92.33 & 93.16 & \textbf{95.26}\\
	Fold 2  & 27.67 & 88.12 & 94.21 & 90.98 & 94.66 & 93.68 & \textbf{95.49}\\
	Fold 3  & 36.32 & 86.32 & 90.90 & 90.38 & 93.83 & 92.41 & \textbf{94.66}\\
	Fold 4  & 35.86 & 87.52 & 93.31 & 91.58 & 94.21 & 94.14 & \textbf{95.34}\\
	Fold 5  & 35.41 & 86.99 & 94.06 & 91.43 & 95.11 & 94.21 & \textbf{95.94}\\
	Fold 6  & 38.87 & 89.17 & 86.92 & 89.62 & 93.61 & 92.26 & \textbf{94.36}\\
	Fold 7  & 30.08 & 83.38 & 93.23 & 89.92 & 93.08 & 93.46 & \textbf{94.96}\\
	Fold 8  & 33.46 & 87.59 & 93.23 & 91.35 & 93.31 & 93.68 & \textbf{95.11}\\
	Fold 9  & 31.95 & 86.92 & 92.78 & 91.43 & 93.38 & 94.14 & \textbf{94.96}\\
	Fold 10  & 36.99 & 84.06 & 92.63 & 91.28 & 93.16 & 93.68 & \textbf{95.19}\\
	\hline\hline
	mean &33.99 & 86.59 & 92.44 & 90.71 & 93.67 & 93.48 &  \textbf{95.13}\\
	std &3.40 & 1.78 & 2.14 & 0.89 & 0.82 & 0.69 & 0.44\\
	\hline
\end{tabular}
\end{table}

\begin{figure}	
\begin{tabular}{@{}c@{}c@{}}
	\includegraphics[width=0.45\textwidth]{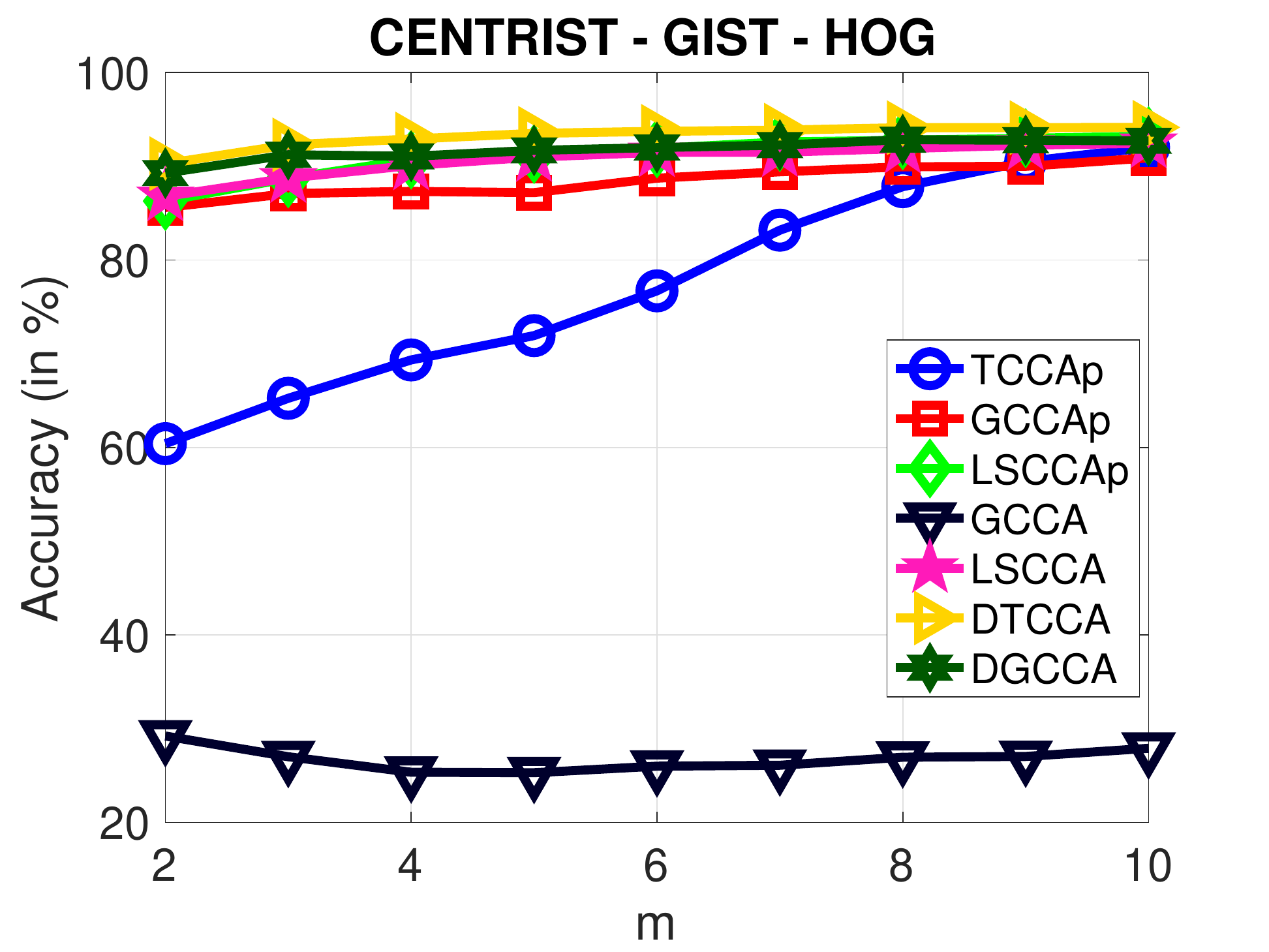} & 
	\includegraphics[width=0.45\textwidth]{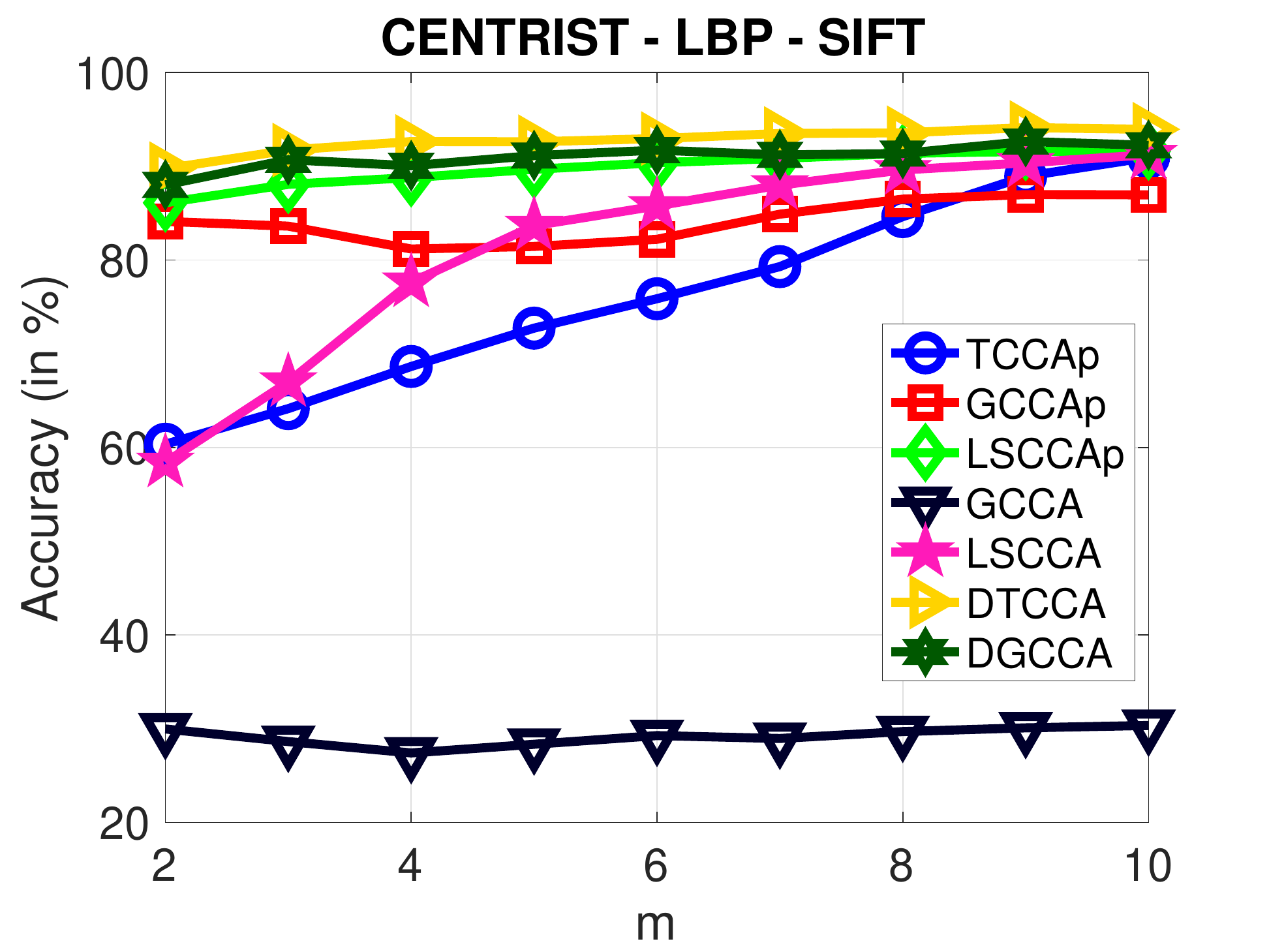} \\
	\includegraphics[width=0.45\textwidth]{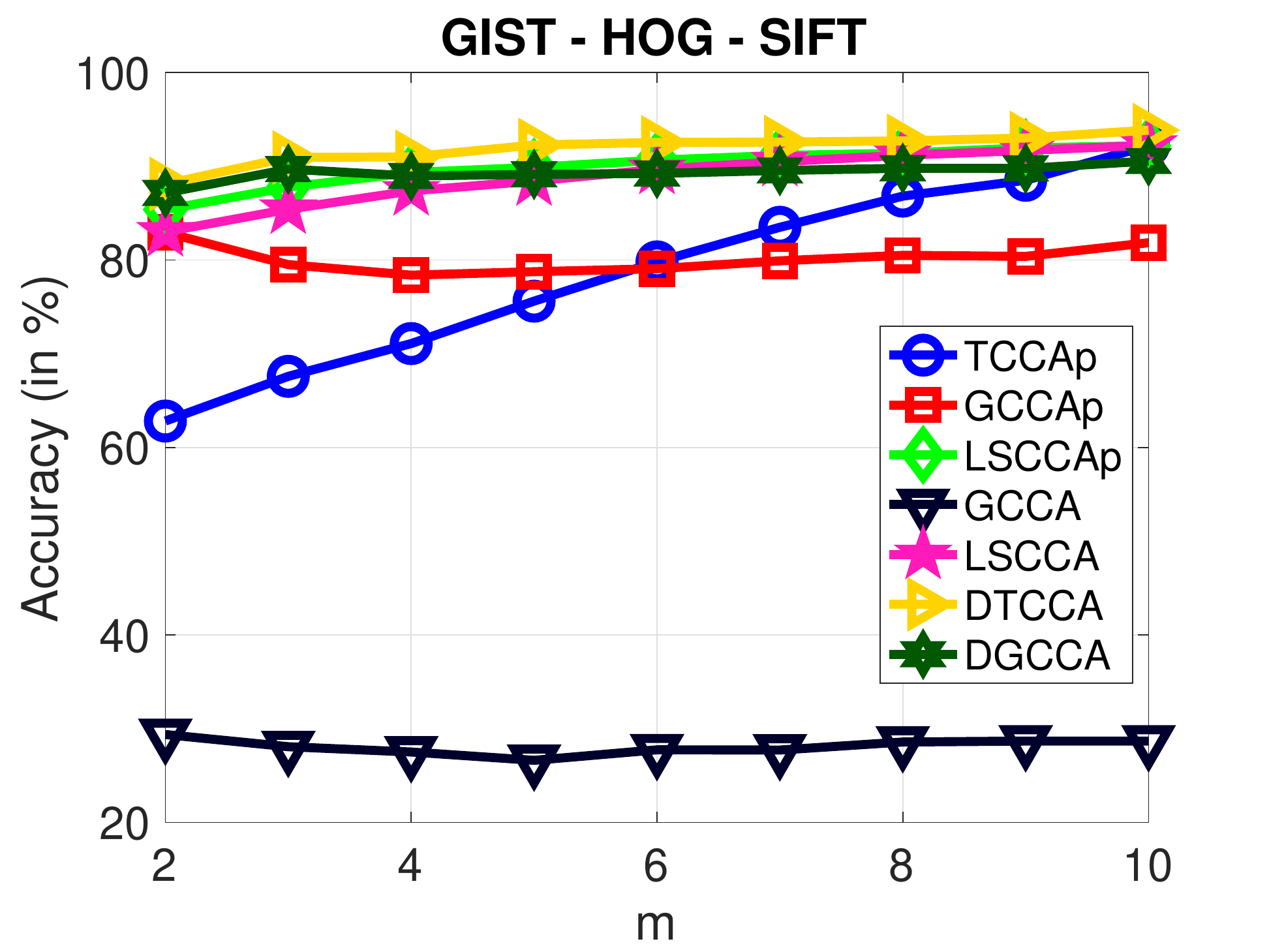} &
	\includegraphics[width=0.45\textwidth]{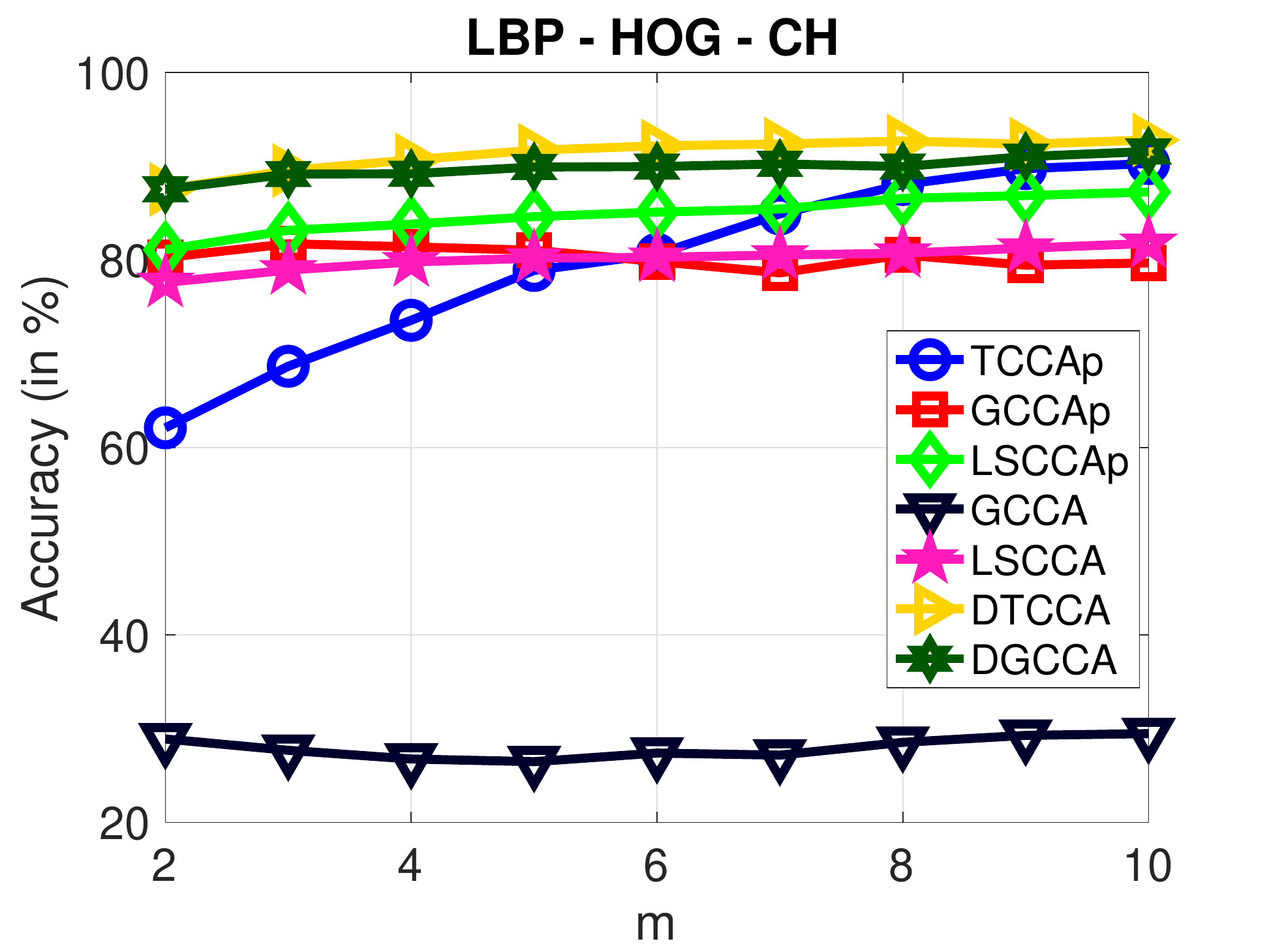} 
\end{tabular}
\caption{The accuracy of seven compared methods on Caltech101 over three views by varying the dimension of the reduced space $m \in \{2,3,\ldots,10\}$.} \label{fig:caltech101-7-dim}
\end{figure}

Following the work \cite{luo2015tensor}, we first concatenate the projected points of all views obtained by CCA variants in the common space   as the final representation for supervised classification problems, and then evaluate the classification performance in terms of accuracy based on linear support vector classifier (SVC) \cite{chang2011libsvm}. We split the data into training and test sets. Projections are learned by compared methods using the training data, and the final accuracy is reported based on the test data. To avoid the learning bias, we report the average accuracy and standard deviations over ten randomly drawn training/test splits and parameter $C$ in SVC is tuned in the range $\{0.1, 1, 10\}$ by repeating experiments for $10$ times.

\begin{table}
\caption{The accuracy of $7$ compared methods over Caltech101 data set by varying the number of views.} \label{tab:caltech101-view}
\centering
\begin{tabular}{@{}lcccc@{}}
	\hline
	& 3 views&4 views&5 views&6 views\\\hline
	GCCA& 31.02 $\pm$ 1.48& 30.54 $\pm$ 0.91& 30.81 $\pm$ 0.67& 31.09\\
	LSCCA& 83.97 $\pm$ 8.97& 87.11 $\pm$ 3.62& 86.92 $\pm$ 2.95& 86.48\\
	TCCA$_p$& 90.69 $\pm$ 1.65& 90.99 $\pm$ 1.12& 91.60 $\pm$ 0.36& 91.56\\
	GCCA$_p$& 85.65 $\pm$ 3.55& 87.66 $\pm$ 3.23& 89.43 $\pm$ 1.52& 90.35\\
	LSCCA$_p$& 90.76 $\pm$ 2.25& 92.08 $\pm$ 1.20& 92.93 $\pm$ 0.57& 93.31\\
	DGCCA& 91.80 $\pm$ 1.07& 92.70 $\pm$ 0.64& 93.28 $\pm$ 0.27& 93.26\\
	DTCCA& \textbf{93.14 $\pm$ 1.14}& \textbf{93.62 $\pm$ 0.78}& \textbf{94.23 $\pm$ 0.41}& \textbf{ 94.83}\\
	\hline
\end{tabular}
\end{table}

\begin{table}
\caption{The accuracy of seven compared methods on Caltech101 data by varying the number of training data with $6$ views.} \label{tab:training_ratio}
\centering
\begin{tabular}{@{}lccccccc@{}}
	\hline
	& 10\%&20\%&30\%&40\%&50\%&60\%&70\%\\
	\hline
	GCCA& 31.09& 24.45& 28.62& 26.26& 26.31& 33.18& 40.92\\
	LSCCA& 86.48& 79.43& 76.17& 77.34& 79.05& 78.04& 79.93\\
	TCCA$_p$& 91.56& 93.71& 95.78& 96.41& 97.51& 97.58& 97.06\\
	GCCA$_p$& 90.35& 93.33& 94.50& 95.40& 95.91& 96.10& 96.14\\
	LSCCA$_p$& 93.31& 95.59& 96.26& 96.74& 97.21& 97.35& 97.22\\
	DGCCA& 93.26& 95.33& 96.49& 96.88& 97.09& 97.80& 97.40\\
	DTCCA& \textbf{94.83}& \textbf{96.06}& \textbf{96.65}& \textbf{97.02}& \textbf{97.36}& \textbf{97.67}& \textbf{97.60}\\
	\hline
\end{tabular}
\end{table}

\begin{landscape}
\begin{table*}
\caption{Mean accuracy and standard deviation of $7$ compared methods on $20$ data sets generated from Caltech101 by choosing all combinations of more than two views over $10$ folders. } \label{tab:caltech101-all}
\centering
\begin{scriptsize}
\begin{tabular}{@{}l@{}|c@{\hspace{0.9em}}c@{\hspace{0.9em}}c@{\hspace{0.9em}}c@{\hspace{0.9em}}c@{\hspace{0.9em}}c@{\hspace{0.9em}}c@{}}
	\hline
	View combinations &GCCA&LSCCA&TCCA$_p$&GCCA$_p$&LSCCA$_p$&DGCCA&DTCCA\\
	\hline\hline
	CENT - GIST - LBP& 32.68 $\pm$ 6.16& 91.91 $\pm$ 1.01& 92.55 $\pm$ 1.04& 91.25 $\pm$ 1.20& 93.14 $\pm$ 1.22& 92.69 $\pm$ 0.78& \textbf{93.89 $\pm$ 0.72}\\
	CENT - GIST - HOG& 29.22 $\pm$ 6.39& 92.45 $\pm$ 0.84& 92.01 $\pm$ 1.81& 90.89 $\pm$ 1.72& 93.21 $\pm$ 1.04& 92.83 $\pm$ 0.97& \textbf{94.15 $\pm$ 0.76}\\
	CENT - GIST - CH& 34.10 $\pm$ 4.34& 83.64 $\pm$ 1.86& 88.53 $\pm$ 2.54& 90.00 $\pm$ 1.08& 91.53 $\pm$ 1.00& 91.73 $\pm$ 0.76& \textbf{92.89 $\pm$ 1.01}\\
	CENT - GIST - SIFT& 33.11 $\pm$ 3.87& 92.56 $\pm$ 0.87& 89.27 $\pm$ 1.32& 88.38 $\pm$ 1.90& 93.11 $\pm$ 0.99& 92.69 $\pm$ 0.96& \textbf{94.33 $\pm$ 0.89}\\
	CENT - LBP - HOG& 33.26 $\pm$ 5.85& 91.35 $\pm$ 1.10& 92.90 $\pm$ 1.09& 89.33 $\pm$ 1.29& 92.40 $\pm$ 0.86& 93.14 $\pm$ 0.62& \textbf{94.29 $\pm$ 0.54}\\
	CENT - LBP - CH& 31.80 $\pm$ 2.73& 84.12 $\pm$ 1.47& 88.75 $\pm$ 3.81& 89.58 $\pm$ 1.07& 91.26 $\pm$ 0.83& 91.12 $\pm$ 0.65& \textbf{91.76 $\pm$ 0.89}\\
	CENT - LBP - SIFT& 30.35 $\pm$ 1.95& 91.23 $\pm$ 1.44& 90.95 $\pm$ 3.88& 86.98 $\pm$ 1.00& 91.62 $\pm$ 1.16& 92.64 $\pm$ 1.12& \textbf{94.12 $\pm$ 1.10}\\
	CENT - HOG - CH& 32.02 $\pm$ 4.66& 84.53 $\pm$ 1.32& 90.12 $\pm$ 1.38& 88.06 $\pm$ 1.57& 90.94 $\pm$ 1.11& 91.96 $\pm$ 0.93& \textbf{92.98 $\pm$ 0.77}\\
	CENT - HOG - SIFT& 29.44 $\pm$ 4.49& 92.03 $\pm$ 1.19& 91.22 $\pm$ 2.61& 85.44 $\pm$ 1.50& 92.04 $\pm$ 1.88& 92.45 $\pm$ 1.04& \textbf{94.38 $\pm$ 0.80}\\
	CENT - CH - SIFT& 30.57 $\pm$ 4.22& 84.64 $\pm$ 1.07& 89.11 $\pm$ 3.31& 83.83 $\pm$ 3.10& 89.60 $\pm$ 0.95& 91.14 $\pm$ 0.68& \textbf{92.17 $\pm$ 0.94}\\
	GIST - LBP - HOG& 31.53 $\pm$ 7.65& 77.87 $\pm$ 3.48& 94.11 $\pm$ 0.65& 85.95 $\pm$ 3.23& 91.95 $\pm$ 1.45& 92.90 $\pm$ 1.35& \textbf{94.37 $\pm$ 0.66}\\
	GIST - LBP - CH& 30.57 $\pm$ 6.26& 82.32 $\pm$ 2.74& 89.03 $\pm$ 2.72& 85.93 $\pm$ 2.59& 90.56 $\pm$ 0.72& 91.83 $\pm$ 0.87& \textbf{92.40 $\pm$ 0.71}\\
	GIST - LBP - SIFT& 30.65 $\pm$ 2.90& 72.08 $\pm$ 8.97& 91.92 $\pm$ 3.27& 84.93 $\pm$ 1.81& 92.74 $\pm$ 1.19& 92.58 $\pm$ 0.76& \textbf{93.98 $\pm$ 1.24}\\
	GIST - HOG - CH& 29.24 $\pm$ 5.00& 82.61 $\pm$ 2.70& 88.79 $\pm$ 2.04& 81.77 $\pm$ 2.28& 87.60 $\pm$ 1.07& 90.39 $\pm$ 1.27& \textbf{92.17 $\pm$ 0.93}\\
	GIST - HOG - SIFT& 29.38 $\pm$ 5.88& 92.25 $\pm$ 0.92& 92.20 $\pm$ 2.83& 82.94 $\pm$ 1.74& 92.19 $\pm$ 1.62& 90.56 $\pm$ 1.64& \textbf{93.85 $\pm$ 0.45}\\
	GIST - CH - SIFT& 29.42 $\pm$ 3.09& 83.41 $\pm$ 2.25& 89.19 $\pm$ 2.75& 81.73 $\pm$ 3.45& 87.98 $\pm$ 1.53& 90.32 $\pm$ 1.07& \textbf{91.54 $\pm$ 1.16}\\
	LBP - HOG - CH& 29.47 $\pm$ 2.93& 81.76 $\pm$ 2.04& 90.29 $\pm$ 1.61& 81.73 $\pm$ 1.77& 87.25 $\pm$ 2.26& 91.57 $\pm$ 0.92& \textbf{92.80 $\pm$ 0.54}\\
	LBP - HOG - SIFT& 31.51 $\pm$ 2.70& 54.26 $\pm$ 0.24& 92.50 $\pm$ 2.97& 83.56 $\pm$ 1.79& 92.39 $\pm$ 1.14& 92.63 $\pm$ 1.25& \textbf{94.15 $\pm$ 0.92}\\
	LBP - CH - SIFT& 31.56 $\pm$ 3.69& 82.19 $\pm$ 1.98& 89.89 $\pm$ 3.08& 80.53 $\pm$ 2.63& 87.52 $\pm$ 1.43& 91.74 $\pm$ 1.02& \textbf{91.75 $\pm$ 1.75}\\
	HOG - CH - SIFT& 30.49 $\pm$ 5.21& 82.29 $\pm$ 1.98& 90.50 $\pm$ 2.68& 80.15 $\pm$ 3.24& 86.18 $\pm$ 2.14& 89.18 $\pm$ 1.28& \textbf{90.91 $\pm$ 0.86}\\
	\hline\hline
	CENT - GIST - LBP - HOG& 30.02 $\pm$ 5.39& 92.49 $\pm$ 0.99& 92.72 $\pm$ 0.91& 91.17 $\pm$ 0.98& 93.41 $\pm$ 1.11& 93.42 $\pm$ 0.60& \textbf{94.55 $\pm$ 0.87}\\
	CENT - GIST - LBP - CH& 30.20 $\pm$ 4.39& 85.08 $\pm$ 1.88& 90.29 $\pm$ 1.83& 91.20 $\pm$ 1.04& 92.65 $\pm$ 1.02& 92.38 $\pm$ 0.89& \textbf{93.09 $\pm$ 0.57}\\
	CENT - GIST - LBP - SIFT& 30.98 $\pm$ 6.21& 92.55 $\pm$ 1.41& 90.15 $\pm$ 3.47& 89.08 $\pm$ 2.37& 93.45 $\pm$ 0.86& 93.38 $\pm$ 0.77& \textbf{94.35 $\pm$ 0.85}\\
	CENT - GIST - HOG - CH& 30.79 $\pm$ 5.52& 84.75 $\pm$ 1.73& 90.02 $\pm$ 2.31& 90.71 $\pm$ 1.07& 92.50 $\pm$ 1.04& 92.48 $\pm$ 0.80& \textbf{93.46 $\pm$ 0.82}\\
	CENT - GIST - HOG - SIFT& 29.26 $\pm$ 3.50& 92.89 $\pm$ 1.01& 88.91 $\pm$ 4.84& 88.84 $\pm$ 1.95& 93.33 $\pm$ 1.32& 93.04 $\pm$ 1.23& \textbf{94.95 $\pm$ 1.03}\\
	CENT - GIST - CH - SIFT& 32.10 $\pm$ 3.86& 85.14 $\pm$ 1.61& 89.63 $\pm$ 2.86& 89.07 $\pm$ 1.31& 92.42 $\pm$ 0.97& 92.48 $\pm$ 0.82& \textbf{93.70 $\pm$ 0.56}\\
	CENT - LBP - HOG - CH& 30.03 $\pm$ 3.06& 85.19 $\pm$ 1.49& 91.74 $\pm$ 1.37& 89.69 $\pm$ 1.55& 92.16 $\pm$ 1.19& 92.71 $\pm$ 0.69& \textbf{93.26 $\pm$ 0.68}\\
	CENT - LBP - HOG - SIFT& 30.02 $\pm$ 6.87& 92.09 $\pm$ 1.08& 91.93 $\pm$ 2.37& 88.12 $\pm$ 2.14& 92.53 $\pm$ 1.46& 93.37 $\pm$ 0.76& \textbf{94.56 $\pm$ 0.36}\\
	CENT - LBP - CH - SIFT& 32.02 $\pm$ 3.84& 85.32 $\pm$ 1.66& 90.48 $\pm$ 3.89& 88.79 $\pm$ 0.89& 91.85 $\pm$ 0.82& 92.10 $\pm$ 0.81& \textbf{92.95 $\pm$ 1.12}\\
	CENT - HOG - CH - SIFT& 29.81 $\pm$ 2.39& 85.51 $\pm$ 1.34& 90.75 $\pm$ 2.61& 88.71 $\pm$ 2.32& 91.44 $\pm$ 1.13& 92.64 $\pm$ 0.87& \textbf{93.82 $\pm$ 1.17}\\
	GIST - LBP - HOG - CH& 29.84 $\pm$ 2.69& 83.74 $\pm$ 1.71& 91.03 $\pm$ 1.76& 85.61 $\pm$ 3.75& 91.55 $\pm$ 0.95& 92.65 $\pm$ 1.18& \textbf{92.98 $\pm$ 0.95}\\
	GIST - LBP - HOG - SIFT& 29.50 $\pm$ 4.26& 89.46 $\pm$ 1.73& 92.81 $\pm$ 2.39& 86.31 $\pm$ 2.29& 93.15 $\pm$ 0.82& 93.38 $\pm$ 0.77& \textbf{94.44 $\pm$ 0.94}\\
	GIST - LBP - CH - SIFT& 31.14 $\pm$ 4.37& 83.91 $\pm$ 2.30& 91.32 $\pm$ 3.77& 85.55 $\pm$ 3.09& 91.48 $\pm$ 1.29& \textbf{92.93 $\pm$ 0.57}& 92.91 $\pm$ 1.20\\
	GIST - HOG - CH - SIFT& 29.62 $\pm$ 2.85& 84.38 $\pm$ 1.93& 91.98 $\pm$ 2.04& 81.80 $\pm$ 2.82& 89.95 $\pm$ 1.28& 90.94 $\pm$ 0.87& \textbf{92.43 $\pm$ 1.43}\\
	LBP - HOG - CH - SIFT& 31.93 $\pm$ 3.46& 84.10 $\pm$ 2.20& 91.09 $\pm$ 3.07& 80.24 $\pm$ 6.90& 89.35 $\pm$ 2.25& 92.65 $\pm$ 0.99& \textbf{92.84 $\pm$ 0.92}\\
	\hline\hline
	CENT - GIST - LBP - HOG - CH& 30.19 $\pm$ 3.96& 86.00 $\pm$ 1.91& 91.35 $\pm$ 1.34& 91.30 $\pm$ 1.27& 93.22 $\pm$ 1.18& 93.32 $\pm$ 1.18& \textbf{93.95 $\pm$ 0.89}\\
	CENT - GIST - LBP - HOG - SIFT& 30.71 $\pm$ 6.44& 92.88 $\pm$ 1.21& 91.59 $\pm$ 3.31& 89.52 $\pm$ 1.74& 93.59 $\pm$ 0.99& 93.79 $\pm$ 0.84& \textbf{95.03 $\pm$ 0.69}\\
	CENT - GIST - LBP - CH - SIFT& 31.76 $\pm$ 4.24& 86.06 $\pm$ 1.73& 91.46 $\pm$ 2.21& 90.26 $\pm$ 1.25& 93.06 $\pm$ 1.01& 93.01 $\pm$ 0.76& \textbf{94.05 $\pm$ 1.11}\\
	CENT - GIST - HOG - CH - SIFT& 31.53 $\pm$ 3.95& 85.78 $\pm$ 1.64& 91.34 $\pm$ 3.09& 89.75 $\pm$ 1.21& 93.17 $\pm$ 1.00& 93.21 $\pm$ 0.79&\textbf{94.32 $\pm$ 0.92}\\
	CENT - LBP - HOG - CH - SIFT& 30.32 $\pm$ 3.53& 85.89 $\pm$ 1.57& 91.56 $\pm$ 1.96& 89.00 $\pm$ 1.26& 92.52 $\pm$ 1.20& 93.14 $\pm$ 1.00& \textbf{94.07 $\pm$ 0.99}\\
	GIST - LBP - HOG - CH - SIFT& 30.38 $\pm$ 3.80& 84.90 $\pm$ 2.00& 92.30 $\pm$ 2.18& 86.77 $\pm$ 2.81& 92.01 $\pm$ 1.07& 93.22 $\pm$ 0.78& \textbf{94.00 $\pm$ 0.60}\\
	\hline\hline
	CENT - GIST - LBP - HOG - CH - SIFT& 31.09 $\pm$ 5.04& 86.48 $\pm$ 1.84& 91.56 $\pm$ 2.53& 90.35 $\pm$ 0.98& 93.31 $\pm$ 1.04& 93.26 $\pm$ 0.66& \textbf{94.83 $\pm$ 0.28}\\
	\hline
\end{tabular}
\end{scriptsize}
\end{table*}
\end{landscape}

\subsection{Image classification} \label{sec:image-classification}
Image data Caltech101 is used for this analysis \cite{fei2007learning}, which is publicly available\footnote{http://www.vision.caltech.edu/Image\_Datasets/Caltech101/}. Caltech101 consists of $8677$ images belonging to $101$ categories.
We apply $6$ different descriptors to extract features for each image, including: 254-d CENTRIST (CENT) \cite{wu2008place}, 512-d GIST \cite{oliva2001modeling},
1180-d LBP \cite{ojala2002multiresolution}, 1008-d histogram of oriented gradient (HOG), 64-d color histogram (CH),
and 1000-d SIFT \cite{lazebnik2006beyond}. 
For classification evaluation, we choose $7$ categories with 1474 images in total by following \cite{deng2015multi}: Faces, Motorbikes, dollar\_bill, garfield, snoopy, stop\_sign, windsor\_chair. 


We first randomly selected $10\%$ images from Caltech101 as the training set and the rest as the testing set by evaluating $7$ compared methods over all $6$ views. Regularization parameter for all CCA methods is set to $10^{-4}$ in order to prevent the singularity of covariance matrix of each view. The selected widths of the hidden layers for the DTCCA-$m$-3 and DGCCA-$m$-3 models are $500$ and $500$ with $m \in [2, 10]$. The Sigmoid activation function is used, and the dropout with ratio equal to $0.1$ is placed in-between each linear layer and its corresponding nonlinear layer except the output layer. The best results of $7$ methods for all $10$ folders over different $m\in [2,10]$ are reported in Table \ref{tab:splits}. We have the following observations: 1) GCCA shows the worst results compared with others; 2) PCA as the preprocessing is helpful to improve CCA classification; 3) nonlinear representation learning using deep networks including DGCCA and DTCCA can significantly improve their base methods GCCA and TCCA, respectively; 4) the proposed DTCCA demonstrates the best results among all compared methods in terms of mean accuracy and the smallest standard deviation, and shows significant better results than TCCA$_p$. These observations imply that the proposed nonlinear extension of TCCA works reasonably well.

\begin{table}
\caption{Accuracy of two methods with layers ranging from two to seven on Caltech101 with $6$ views. `-' is for the failure of the training process.}
\label{tab:networks}
\centering
\begin{tabular}{@{}lcccccc@{}}
	\hline
	Layers	& 2 & 3 & 4 & 5 & 6 & 7\\
	\hline
	DGCCA (sigmoid) &  93.46& 92.63 & 93.31 & 90.00 & 56.39 & -\\
	DTCCA (sigmoid) & 95.64& 95.04 & 95.41 & 92.11 & 81.58 & -\\
	DGCCA (tanh) & 92.33 & 93.01 & 92.63 & 92.11 & 91.88 & 92.48\\
	DTCCA (tanh) & 94.29 & 94.44 & 94.51 & 94.74 & 95.11 & 95.11\\
	\hline
\end{tabular}
\end{table}

\begin{landscape}
\begin{table*}
\caption{Mean accuracy and standard deviation of $7$ compared methods on $20$ data sets generated from Mfeat data by choosing all combinations of more than two views over $10$ folders. } \label{tab:mfeat-all}
\centering
\begin{scriptsize}
\begin{tabular}{@{}l|ccccccc@{}}
	\hline
	View combinations&GCCA&LSCCA&TCCA$_p$&GCCA$_p$&LSCCA$_p$&DGCCA&DTCCA\\
	\hline\hline
	fac - fou - kar& 26.18 $\pm$ 4.01& 62.80 $\pm$ 3.02& 83.34 $\pm$ 2.83& 79.60 $\pm$ 2.33& 84.73 $\pm$ 2.14& \textbf{85.82 $\pm$ 1.83}& 85.61 $\pm$ 1.10\\
	fac - fou - mor& 39.72 $\pm$ 3.63& 62.05 $\pm$ 4.01& 86.96 $\pm$ 2.81& 83.23 $\pm$ 1.68& 85.39 $\pm$ 1.43& 92.51 $\pm$ 1.17& \textbf{92.62 $\pm$ 0.99}\\
	fac - fou - pix& 13.58 $\pm$ 1.63& 69.15 $\pm$ 3.44& 83.45 $\pm$ 1.58& 81.84 $\pm$ 2.01& 85.73 $\pm$ 1.98& 86.75 $\pm$ 1.47& \textbf{88.91 $\pm$ 1.08}\\
	fac - fou - zer& 30.82 $\pm$ 5.15& 64.69 $\pm$ 2.58& 79.61 $\pm$ 3.21& 78.92 $\pm$ 2.17& 82.89 $\pm$ 1.66& 86.59 $\pm$ 1.70& \textbf{87.67 $\pm$ 0.94}\\
	fac - kar - mor& 37.52 $\pm$ 6.46& 69.84 $\pm$ 5.68& 87.05 $\pm$ 1.48& 84.90 $\pm$ 2.62& 88.30 $\pm$ 2.62& 91.39 $\pm$ 0.67& \textbf{92.69 $\pm$ 1.02}\\
	fac - kar - pix& 14.26 $\pm$ 1.09& 80.84 $\pm$ 2.34& 77.73 $\pm$ 1.86& 79.18 $\pm$ 2.35& 85.43 $\pm$ 1.34& 82.08 $\pm$ 1.70& \textbf{87.82 $\pm$ 1.70}\\
	fac - kar - zer& 31.72 $\pm$ 2.74& 65.68 $\pm$ 2.48& 79.03 $\pm$ 2.57& 73.70 $\pm$ 2.77& 77.35 $\pm$ 3.03& 83.12 $\pm$ 1.56& \textbf{86.96 $\pm$ 0.79}\\
	fac - mor - pix& 12.52 $\pm$ 2.05& 55.00 $\pm$ 6.94& 88.89 $\pm$ 2.11& 83.78 $\pm$ 2.68& 86.84 $\pm$ 2.64& 92.08 $\pm$ 0.61& \textbf{94.14 $\pm$ 0.53}\\
	fac - mor - zer& 39.04 $\pm$ 2.79& 67.52 $\pm$ 2.59& 84.07 $\pm$ 2.25& 80.61 $\pm$ 1.90& 82.59 $\pm$ 1.93& 91.32 $\pm$ 0.67& \textbf{93.21 $\pm$ 0.89}\\
	fac - pix - zer& 13.48 $\pm$ 1.82& 61.83 $\pm$ 2.26& 80.76 $\pm$ 3.92& 71.18 $\pm$ 2.68& 75.17 $\pm$ 2.37& 85.64 $\pm$ 1.42& \textbf{89.47 $\pm$ 1.05}\\
	fou - kar - mor& 69.55 $\pm$ 2.25& 73.95 $\pm$ 2.22& 85.77 $\pm$ 1.34& 80.02 $\pm$ 1.62& 83.03 $\pm$ 1.70& \textbf{87.61 $\pm$ 1.62}& 87.51 $\pm$ 1.24\\
	fou - kar - pix& 13.61 $\pm$ 1.77& 60.19 $\pm$ 3.85& 82.46 $\pm$ 2.85& 73.81 $\pm$ 3.39& 80.16 $\pm$ 3.11& 83.52 $\pm$ 1.23& \textbf{84.18 $\pm$ 1.63}\\
	fou - kar - zer& 63.83 $\pm$ 1.70& 74.33 $\pm$ 1.70& 77.96 $\pm$ 3.50& 75.72 $\pm$ 3.37& 80.24 $\pm$ 2.90& \textbf{81.23 $\pm$ 1.15}& 80.28 $\pm$ 1.83\\
	fou - mor - pix& 11.89 $\pm$ 1.52& 62.20 $\pm$ 3.71& 86.96 $\pm$ 2.36& 80.19 $\pm$ 2.16& 82.81 $\pm$ 1.56& 90.98 $\pm$ 2.05& \textbf{92.43 $\pm$ 0.72}\\
	fou - mor - zer& 71.16 $\pm$ 3.02& 74.58 $\pm$ 2.33& 79.64 $\pm$ 1.25& 78.03 $\pm$ 1.74& 80.29 $\pm$ 1.33& \textbf{81.18 $\pm$ 0.77}& 79.75 $\pm$ 0.70\\
	fou - pix - zer& 12.30 $\pm$ 1.53& 61.30 $\pm$ 2.34& 78.88 $\pm$ 2.56& 76.77 $\pm$ 1.68& 80.59 $\pm$ 1.60& 85.26 $\pm$ 1.59& \textbf{85.58 $\pm$ 0.98}\\
	kar - mor - pix& 13.37 $\pm$ 1.43& 69.11 $\pm$ 5.24& 85.67 $\pm$ 2.16& 79.41 $\pm$ 2.89& 82.88 $\pm$ 2.79& 89.68 $\pm$ 1.34& \textbf{92.80 $\pm$ 0.72}\\
	kar - mor - zer& 72.71 $\pm$ 2.77& 76.68 $\pm$ 2.67& 83.95 $\pm$ 2.12& 79.94 $\pm$ 2.01& 82.87 $\pm$ 1.83& 86.91 $\pm$ 1.85& \textbf{89.21 $\pm$ 1.10}\\
	kar - pix - zer& 15.10 $\pm$ 2.29& 65.11 $\pm$ 2.81& 75.29 $\pm$ 2.12& 65.22 $\pm$ 1.88& 71.21 $\pm$ 2.60& 83.15 $\pm$ 1.58& \textbf{86.30 $\pm$ 1.76}\\
	mor - pix - zer& 13.87 $\pm$ 2.51& 63.39 $\pm$ 2.06& 85.01 $\pm$ 1.47& 79.08 $\pm$ 1.55& 81.79 $\pm$ 1.75& 89.73 $\pm$ 1.00& \textbf{92.81 $\pm$ 0.65}\\
	\hline\hline
	fac - fou - kar - mor& 41.02 $\pm$ 5.67& 73.00 $\pm$ 1.91& 86.87 $\pm$ 2.25& 83.64 $\pm$ 2.44& 88.29 $\pm$ 1.78& 91.88 $\pm$ 0.64& \textbf{92.53 $\pm$ 0.59}\\
	fac - fou - kar - pix& 13.06 $\pm$ 1.69& 64.89 $\pm$ 2.95& 79.75 $\pm$ 2.39& 77.73 $\pm$ 2.90& 86.21 $\pm$ 2.07& 84.59 $\pm$ 1.79& \textbf{88.62 $\pm$ 0.77}\\
	fac - fou - kar - zer& 32.44 $\pm$ 3.98& 74.89 $\pm$ 1.14& 77.94 $\pm$ 3.10& 79.57 $\pm$ 1.88& 85.60 $\pm$ 1.90& 85.85 $\pm$ 1.13& \textbf{87.11 $\pm$ 1.20}\\
	fac - fou - mor - pix& 11.77 $\pm$ 2.02& 60.92 $\pm$ 3.33& 87.84 $\pm$ 2.87& 83.77 $\pm$ 2.40& 88.58 $\pm$ 1.64& 92.41 $\pm$ 1.33& \textbf{93.87 $\pm$ 0.44}\\
	fac - fou - mor - zer& 39.43 $\pm$ 5.12& 72.76 $\pm$ 0.91& 83.62 $\pm$ 4.00& 81.29 $\pm$ 1.46& 85.07 $\pm$ 1.43& 91.60 $\pm$ 0.99& \textbf{93.28 $\pm$ 0.80}\\
	fac - fou - pix - zer& 12.11 $\pm$ 1.53& 64.72 $\pm$ 2.35& 79.04 $\pm$ 3.27& 79.40 $\pm$ 1.86& 85.56 $\pm$ 1.66& 86.59 $\pm$ 1.03& \textbf{90.60 $\pm$ 1.00}\\
	fac - kar - mor - pix& 14.02 $\pm$ 0.75& 69.27 $\pm$ 5.43& 86.46 $\pm$ 1.50& 84.68 $\pm$ 2.95& 90.72 $\pm$ 1.69& 91.08 $\pm$ 1.15& \textbf{94.50 $\pm$ 0.66}\\
	fac - kar - mor - zer& 39.37 $\pm$ 3.92& 74.68 $\pm$ 2.72& 86.16 $\pm$ 2.33& 81.16 $\pm$ 1.58& 85.82 $\pm$ 1.91& 90.22 $\pm$ 1.24& \textbf{93.62 $\pm$ 0.86}\\
	fac - kar - pix - zer& 15.45 $\pm$ 2.49& 66.48 $\pm$ 2.36& 78.28 $\pm$ 2.46& 71.48 $\pm$ 2.87& 78.47 $\pm$ 2.91& 83.07 $\pm$ 1.80& \textbf{89.75 $\pm$ 0.53}\\
	fac - mor - pix - zer& 12.87 $\pm$ 2.73& 64.54 $\pm$ 2.20& 85.23 $\pm$ 2.72& 80.33 $\pm$ 1.84& 84.61 $\pm$ 1.81& 91.82 $\pm$ 1.20& \textbf{94.59 $\pm$ 0.59}\\
	fou - kar - mor - pix& 14.31 $\pm$ 0.73& 71.92 $\pm$ 1.84& 85.73 $\pm$ 1.84& 79.48 $\pm$ 2.17& 85.25 $\pm$ 1.64& 90.31 $\pm$ 1.05& \textbf{92.19 $\pm$ 1.05}\\
	fou - kar - mor - zer& 71.01 $\pm$ 2.79& 79.72 $\pm$ 1.60& 84.27 $\pm$ 2.80& 80.73 $\pm$ 2.43& 85.03 $\pm$ 2.16& 87.11 $\pm$ 1.54& \textbf{88.62 $\pm$ 1.60}\\
	fou - kar - pix - zer& 15.40 $\pm$ 2.21& 74.01 $\pm$ 1.72& 77.18 $\pm$ 3.40& 76.07 $\pm$ 1.89& 82.77 $\pm$ 1.91& 84.28 $\pm$ 1.59& \textbf{86.73 $\pm$ 1.72}\\
	fou - mor - pix - zer& 13.12 $\pm$ 1.75& 70.47 $\pm$ 2.37& 85.17 $\pm$ 2.45& 79.86 $\pm$ 1.81& 84.96 $\pm$ 1.58& 90.36 $\pm$ 1.68& \textbf{92.48 $\pm$ 0.56}\\
	kar - mor - pix - zer& 13.86 $\pm$ 2.49& 73.28 $\pm$ 2.26& 85.22 $\pm$ 2.40& 77.94 $\pm$ 2.27& 83.24 $\pm$ 1.58& 89.97 $\pm$ 1.69& \textbf{93.39 $\pm$ 0.86}\\
	\hline\hline
	fac - fou - kar - mor - pix& 14.38 $\pm$ 1.34& 72.72 $\pm$ 1.90& 88.02 $\pm$ 2.76& 81.93 $\pm$ 3.16& 89.27 $\pm$ 1.99& 91.76 $\pm$ 0.62& \textbf{93.87 $\pm$ 0.63}\\
	fac - fou - kar - mor - zer& 39.92 $\pm$ 5.13& 79.48 $\pm$ 1.11& 85.76 $\pm$ 1.85& 81.93 $\pm$ 2.17& 88.03 $\pm$ 1.87& 91.72 $\pm$ 0.90& \textbf{93.23 $\pm$ 0.68}\\
	fac - fou - kar - pix - zer& 15.30 $\pm$ 2.36& 75.42 $\pm$ 1.18& 79.94 $\pm$ 2.41& 78.84 $\pm$ 1.30& 86.75 $\pm$ 1.82& 86.34 $\pm$ 1.75& \textbf{89.72 $\pm$ 1.19}\\
	fac - fou - mor - pix - zer& 13.14 $\pm$ 2.35& 71.19 $\pm$ 1.01& 86.69 $\pm$ 3.08& 81.73 $\pm$ 2.70& 87.39 $\pm$ 1.81& 91.99 $\pm$ 1.00& \textbf{94.09 $\pm$ 0.98}\\
	fac - kar - mor - pix - zer& 13.02 $\pm$ 1.98& 73.73 $\pm$ 2.06& 86.17 $\pm$ 1.89& 79.34 $\pm$ 2.03& 86.46 $\pm$ 1.60& 90.73 $\pm$ 0.60& \textbf{94.18 $\pm$ 0.76}\\
	fou - kar - mor - pix - zer& 15.36 $\pm$ 2.22& 78.29 $\pm$ 1.45& 85.11 $\pm$ 2.67& 80.42 $\pm$ 2.15& 86.49 $\pm$ 2.07& 90.14 $\pm$ 1.24& \textbf{92.92 $\pm$ 0.88}\\
	\hline\hline
	fac - fou - kar - mor - pix - zer& 14.94 $\pm$ 2.33& 79.12 $\pm$ 1.39& 87.56 $\pm$ 2.56& 80.73 $\pm$ 2.59& 89.03 $\pm$ 1.79& 91.86 $\pm$ 1.03& \textbf{94.21 $\pm$ 0.61}\\
	\hline
\end{tabular}
\end{scriptsize}
\end{table*}		
\end{landscape}

We further explore the sensitivity of DTCCA in terms of different dimension of the common space, the number of views,  and various training ratios. Since TCCA can naturally incorporate the high-order canonical correlation, we expect that DTCCA or TCCA can perform consistently well in regardless of the number of views. Following the above setting, we varied the number of views from $3$ to $6$ with $10\%$ training data. We first investigate the impact of dimensions by varying $m \in [2, 10]$. The results are shown in Fig. \ref{fig:caltech101-7-dim} for some  combinations of views. All methods except GCCA demonstrates better results when $m$ increases, while DTCCA demonstrates consistently the best over all these dimensions. We observed the same trends for other combinations. Due to the space limitation, we will not report the results for every combination. To investigate the impact of the number of views,
for the specific view, we report the averaged accuracy over all combinations of views, where the accuracy of each combination is again obtained based on the $10$ folders. Results are shown in Table \ref{tab:caltech101-view} for the number of views varying from $3$ to $6$. As the number of views increases, all methods demonstrate improved accuracy, so learning with multiple views becomes important. Also, our proposed DTCCA not only significantly outperforms TCCA and other linear models, but also better than DGCCA with same deep network architecture. Moreover, we evaluate all methods by varying the number of training data with ratio from 10\% to 70\%. Results in Table \ref{tab:training_ratio} demonstrates that DTCCA performs significantly better than others for small amount of training data, and the accuracies obtained by DGCCA, DGCCA, GCCA$_p$ and TCCA$_p$ converge to a similar value when enough training data becomes available.  To show the performance of all combined views, the mean and standard deviation of accuracies of compared methods over $10$ folders for each combination of views are shown in Table \ref{tab:caltech101-all}.

To determine the impact of the number of layers in the deep networks based models such as DGCCA and DTCCA, we conduct an experiment in which we increase the number of layers from two to seven. The width of each hidden layer is set to be $500$. $10\%$ training data split is used for this experiment with $m=10$ for one of $10$ folds. Table \ref{tab:networks} gives the accuracy on the first fold by varying the number of layers from $2$ to $7$ with nonlinear activation function either Sigmoid or Tanh. We have the following observations: 1) Sigmoid function can obtain good results by using the network of $4$ layers, but it becomes worse when the number of layers increases, which is because of the drawback of Sigmoid with zero gradient for deep depth of network; 2) the activation function Tanh does not have this issue so the accuracy continues increasing until 7 layers and the performance reaches saturation.

\subsection{Handwritten numeral recognition}

Multiple features (Mfeat) data set consists of features of handwritten numerals (`0'--`9') extracted from a collection of Dutch utility maps\footnote{https://archive.ics.uci.edu/ml/data sets/Multiple+Features}   \cite{Dua:2019}.  200 patterns per class (for a total of 2,000 patterns) have been digitized in binary images. These digits are represented in terms of the following six feature sets: 216-d profile correlations (fac), 76-d Fourier coefficients of the character shapes (fou),  64-d Karhunen-Love coefficients (kar), 6-d morphological features (mor),
240-d pixel averages in $2 \times 3$ windows (pix), and 47-d Zernike moments (zer). As a result, there are 6 views in total.

\begin{figure}
\begin{tabular}{@{}c@{}c@{}}
	\includegraphics[width=0.45\textwidth]{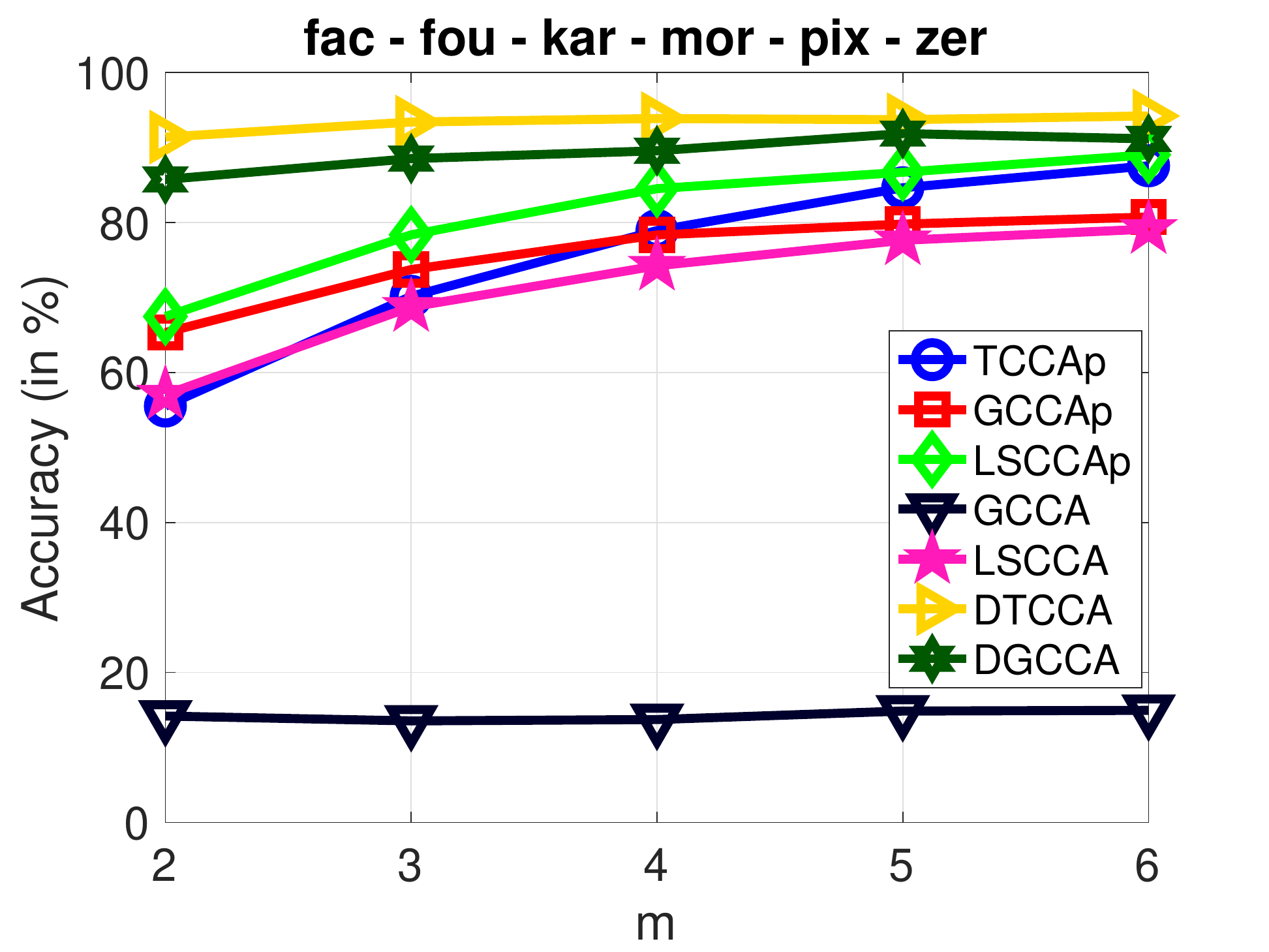} & 
	\includegraphics[width=0.45\textwidth]{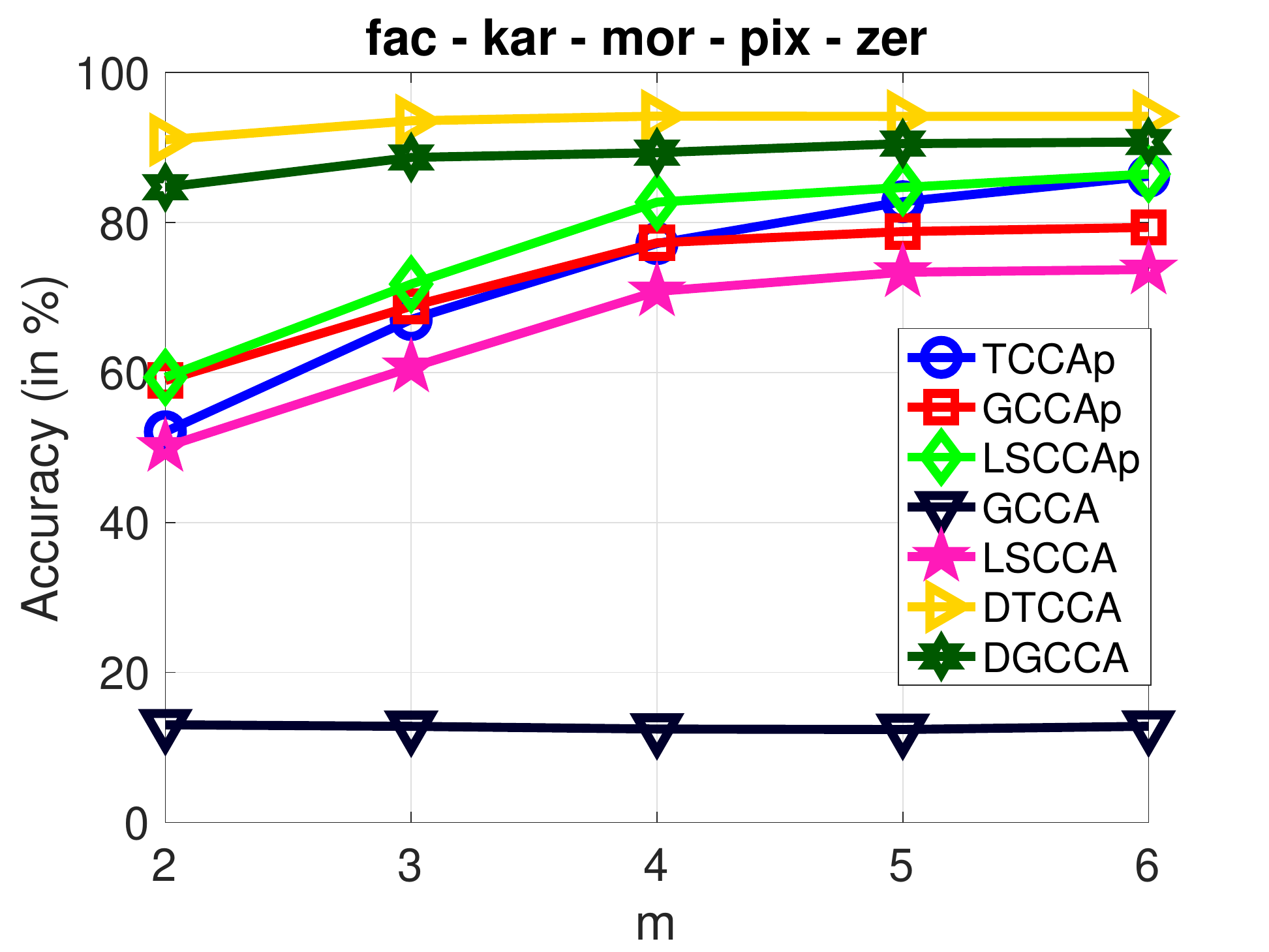} \\
	\includegraphics[width=0.45\textwidth]{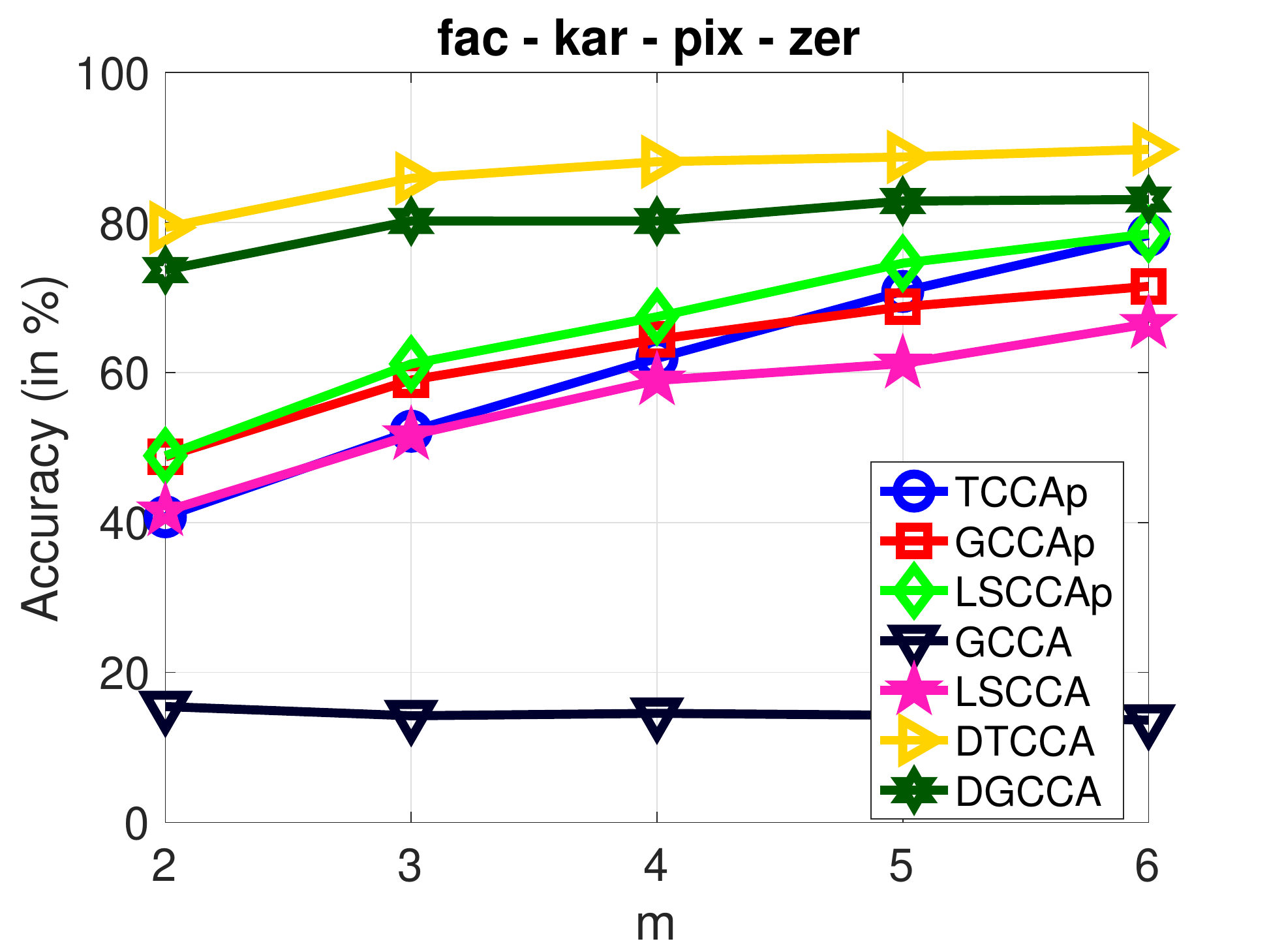} &
	\includegraphics[width=0.45\textwidth]{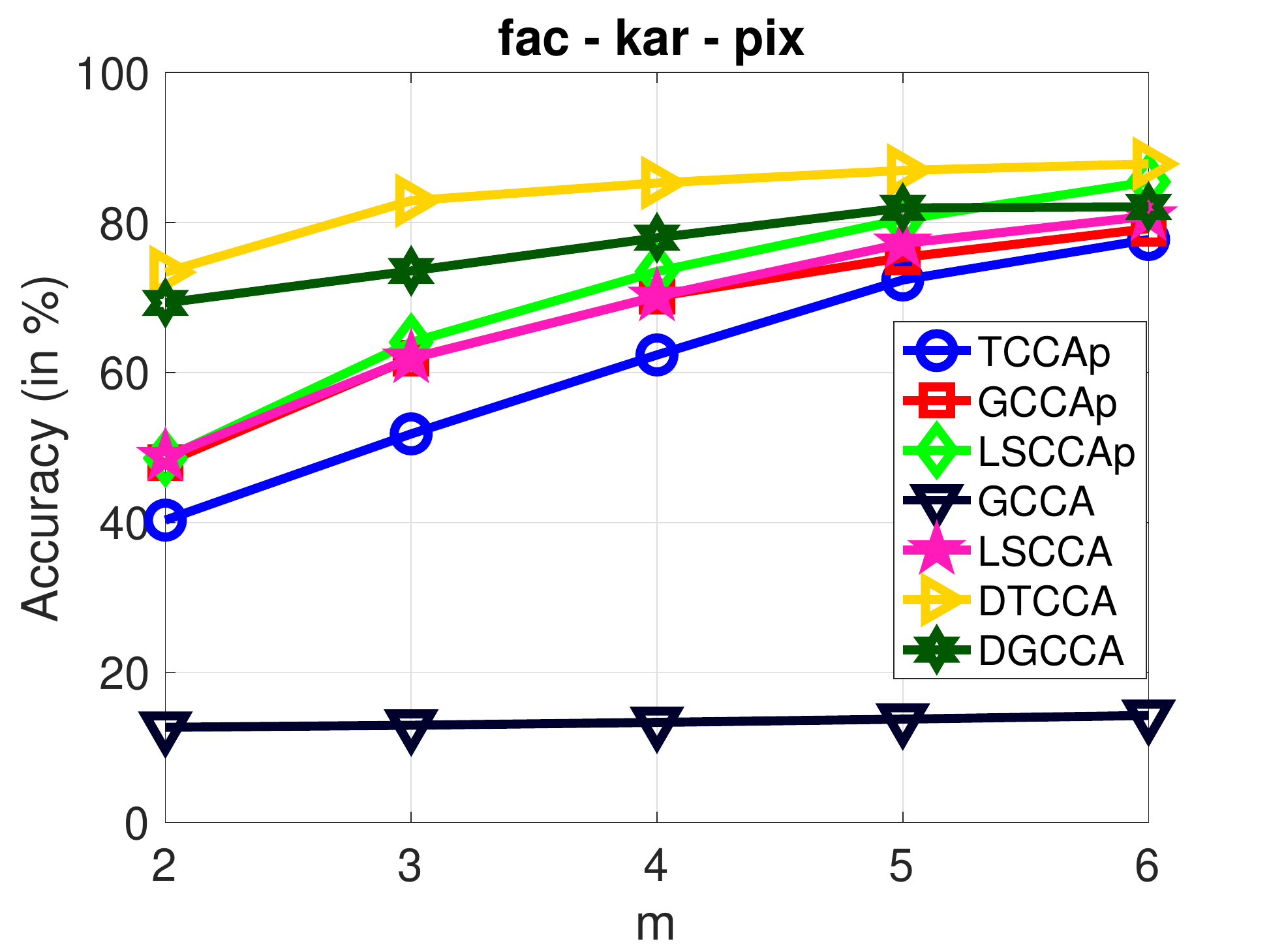} 
\end{tabular}
\caption{The accuracy of seven compared methods on Mfeat data by varying the dimension of the reduced space $m \in \{2,3,\ldots,10\}$.} \label{fig:mfeat-dim}
\end{figure}

The same experimental setting as in Section \ref{sec:image-classification} is used for Mfeat data. Results presented in Table \ref{tab:mfeat-all} shows the classification performance of $7$ compared methods on all combinations of views with $10\%$ as training data and the rest as testing data. The impacts of the compared methods in terms of varied combinations of views, the reduced dimensions and the varied training ratio are shown in Table \ref{tab:mfeat-view}, Fig. \ref{fig:mfeat-dim}, Table \ref{tab:mfeat-train} respectively. These observations are consistent with what we have observed for Caltech101.

\begin{table}
\caption{The accuracy of $7$ compared methods over Mfeat data set by varying the number of views.} \label{tab:mfeat-view}
\centering
\begin{tabular}{@{}lcccc@{}} 
	\hline
	& 3 views&4 views&5 views&6 views\\
	\hline
	GCCA& 30.81 $\pm$ 22.04& 23.95 $\pm$ 17.20& 18.52 $\pm$ 10.53& 14.94\\
	LSCCA& 67.01 $\pm$ 6.49& 70.37 $\pm$ 5.10& 75.14 $\pm$ 3.23& 79.12\\
	TCCA$_p$& 82.62 $\pm$ 3.80& 83.25 $\pm$ 3.70& 85.28 $\pm$ 2.80& 87.56\\
	GCCA$_p$& 78.26 $\pm$ 4.60& 79.81 $\pm$ 3.28& 80.70 $\pm$ 1.37& 80.73\\
	LSCCA$_p$& 82.01 $\pm$ 4.03& 85.35 $\pm$ 2.78& 87.40 $\pm$ 1.10& 89.03\\
	DGCCA& 86.83 $\pm$ 3.73& 88.74 $\pm$ 3.16& 90.45 $\pm$ 2.13& 91.86\\
	DTCCA& \textbf{88.50 $\pm$ 4.18}& \textbf{91.46 $\pm$ 2.68}& \textbf{93.00 $\pm$ 1.68}& \textbf{94.21}\\
	\hline
\end{tabular}
\end{table}

\begin{table}
\caption{The accuracy of seven compared methods on Mfeat data by varying the number of training data with $6$ views.} \label{tab:mfeat-train}
\centering
\begin{tabular}{@{}lccccccc@{}}
	\hline
	& 10\%&20\%&30\%&40\%&50\%&60\%&70\%\\
	\hline
	GCCA& 14.94& 75.68& 84.96& 87.69& 88.74& 89.25& 89.31\\
	LSCCA& 79.12& 86.88& 88.59& 89.69& 90.12& 89.92& 90.21\\
	TCCA$_p$& 87.56& 91.13& 92.97& 94.57& 94.56& 95.15& 95.21\\
	GCCA$_p$& 80.73& 86.74& 89.97& 90.58& 91.61& 92.53& 92.39\\
	LSCCA$_p$& 89.03& 90.00& 91.46& 91.70& 92.01& 92.58& 92.79\\
	DGCCA& 91.86& 94.09& 95.69& 96.47& 96.89& 97.23& \textbf{97.72}\\
	DTCCA& \textbf{94.21}& \textbf{95.69}& \textbf{96.54}& \textbf{96.78}& \textbf{97.14}& \textbf{97.56}& 97.59\\
	\hline
\end{tabular}
\end{table}

\begin{landscape}
\begin{table*}
\caption{Mean accuracy and standard deviation of $7$ compared methods on $20$ data sets generated from Scene15 data by choosing all combinations of more than two views over $10$ folders. } \label{tab:scene15-all}
\centering
\begin{scriptsize}
\begin{tabular}{@{}l@{}|ccccccc@{}}
	\hline
	View combinations &GCCA&LSCCA&TCCA$_p$&GCCA$_p$&LSCCA$_p$&DGCCA&DTCCA\\
	\hline\hline
	CENTRIST - GIST - LBP& 8.03 $\pm$ 0.73& 50.83 $\pm$ 1.86& 60.13 $\pm$ 1.90& 58.39 $\pm$ 1.86& 62.34 $\pm$ 1.45& \textbf{64.79 $\pm$ 0.95}& 64.49 $\pm$ 1.08\\
	CENTRIST - GIST - HOG& 8.81 $\pm$ 0.69& 35.26 $\pm$ 1.92& 56.76 $\pm$ 2.75& 56.05 $\pm$ 1.43& 59.18 $\pm$ 1.28& \textbf{64.57 $\pm$ 1.94}& 64.11 $\pm$ 0.42\\
	CENTRIST - GIST - SIFT& 9.64 $\pm$ 0.58& 50.48 $\pm$ 2.19& 68.70 $\pm$ 1.87& 56.34 $\pm$ 2.34& 61.23 $\pm$ 1.96& 69.64 $\pm$ 1.35& \textbf{70.97 $\pm$ 1.46}\\
	CENTRIST - LBP - HOG& 8.15 $\pm$ 0.64& 42.62 $\pm$ 1.78& 56.82 $\pm$ 1.42& 54.06 $\pm$ 1.49& 57.20 $\pm$ 1.43& 61.35 $\pm$ 1.14& \textbf{62.55 $\pm$ 1.37}\\
	CENTRIST - LBP - SIFT& 8.46 $\pm$ 1.09& 58.72 $\pm$ 2.08& 64.87 $\pm$ 2.96& 50.35 $\pm$ 1.93& 56.50 $\pm$ 1.78& 65.22 $\pm$ 1.46& \textbf{69.87 $\pm$ 1.32}\\
	CENTRIST - HOG - SIFT& 8.70 $\pm$ 0.87& 43.50 $\pm$ 2.03& 65.21 $\pm$ 2.79& 51.32 $\pm$ 2.28& 55.64 $\pm$ 1.58& 66.66 $\pm$ 2.09& \textbf{69.64 $\pm$ 1.56}\\
	GIST - LBP - HOG& 8.31 $\pm$ 0.65& 43.23 $\pm$ 1.22& 54.60 $\pm$ 1.78& 46.08 $\pm$ 1.54& 48.76 $\pm$ 1.66& \textbf{59.92 $\pm$ 0.74}& 59.58 $\pm$ 0.99\\
	GIST - LBP - SIFT& 8.72 $\pm$ 0.57& 55.43 $\pm$ 1.64& 67.37 $\pm$ 3.24& 50.37 $\pm$ 1.83& 55.56 $\pm$ 1.91& 63.59 $\pm$ 2.14& \textbf{66.45 $\pm$ 1.51}\\
	GIST - HOG - SIFT& 9.79 $\pm$ 0.62& 45.03 $\pm$ 1.07& 64.87 $\pm$ 3.67& 44.97 $\pm$ 2.19& 49.21 $\pm$ 2.01& 62.34 $\pm$ 1.10& \textbf{65.19 $\pm$ 1.08}\\
	LBP - HOG - SIFT& 8.28 $\pm$ 0.98& 50.91 $\pm$ 0.83& 64.25 $\pm$ 2.31& 44.76 $\pm$ 1.21& 49.58 $\pm$ 1.73& 61.50 $\pm$ 1.56& \textbf{65.18 $\pm$ 1.62}\\
	\hline\hline
	CENTRIST - GIST - LBP - HOG& 8.56 $\pm$ 0.96& 37.90 $\pm$ 1.71& 55.45 $\pm$ 1.33& 55.28 $\pm$ 1.90& 61.37 $\pm$ 1.09& \textbf{63.60 $\pm$ 1.53}& 62.92 $\pm$ 0.79\\
	CENTRIST - GIST - LBP - SIFT& 9.20 $\pm$ 0.70& 53.61 $\pm$ 2.00& 66.24 $\pm$ 2.32& 54.25 $\pm$ 2.85& 63.14 $\pm$ 2.01& 65.76 $\pm$ 1.67& \textbf{68.01 $\pm$ 1.98}\\
	CENTRIST - GIST - HOG - SIFT& 9.30 $\pm$ 0.81& 38.93 $\pm$ 2.75& 64.08 $\pm$ 2.38& 53.60 $\pm$ 2.13& 60.38 $\pm$ 1.71& 66.73 $\pm$ 1.45& \textbf{66.96 $\pm$ 1.18}\\
	CENTRIST - LBP - HOG - SIFT& 8.79 $\pm$ 0.99& 46.26 $\pm$ 1.61& 64.94 $\pm$ 1.92& 50.33 $\pm$ 1.85& 59.04 $\pm$ 1.57& 63.46 $\pm$ 1.79& \textbf{66.24 $\pm$ 1.85}\\
	GIST - LBP - HOG - SIFT& 9.24 $\pm$ 0.76& 46.91 $\pm$ 1.07& 62.65 $\pm$ 2.57& 45.24 $\pm$ 1.44& 51.47 $\pm$ 1.23& 62.88 $\pm$ 1.82& \textbf{64.73 $\pm$ 1.49}\\
	\hline\hline
	CENTRIST - GIST - LBP - HOG - SIFT& 9.44 $\pm$ 1.00& 41.07 $\pm$ 2.08& 64.45 $\pm$ 1.75& 51.27 $\pm$ 1.72& 62.72 $\pm$ 1.41& 64.71 $\pm$ 1.08& \textbf{67.33 $\pm$ 1.43}\\
	\hline
\end{tabular}
\end{scriptsize}
\end{table*}		
\end{landscape}

\subsection{Scene classification}
\begin{figure}
\begin{tabular}{@{}c@{}c@{}}
	\includegraphics[width=0.45\textwidth]{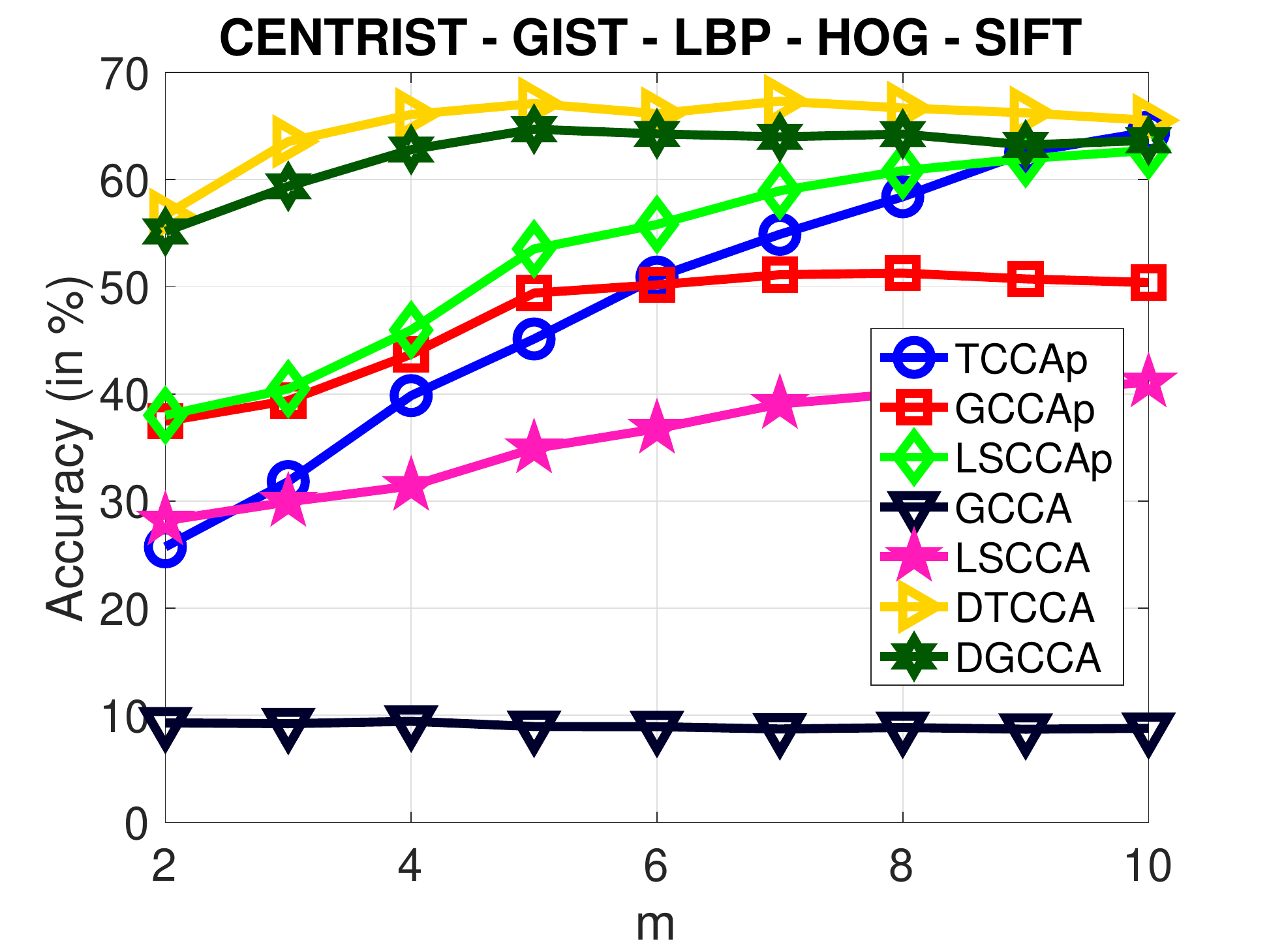} & 
	\includegraphics[width=0.45\textwidth]{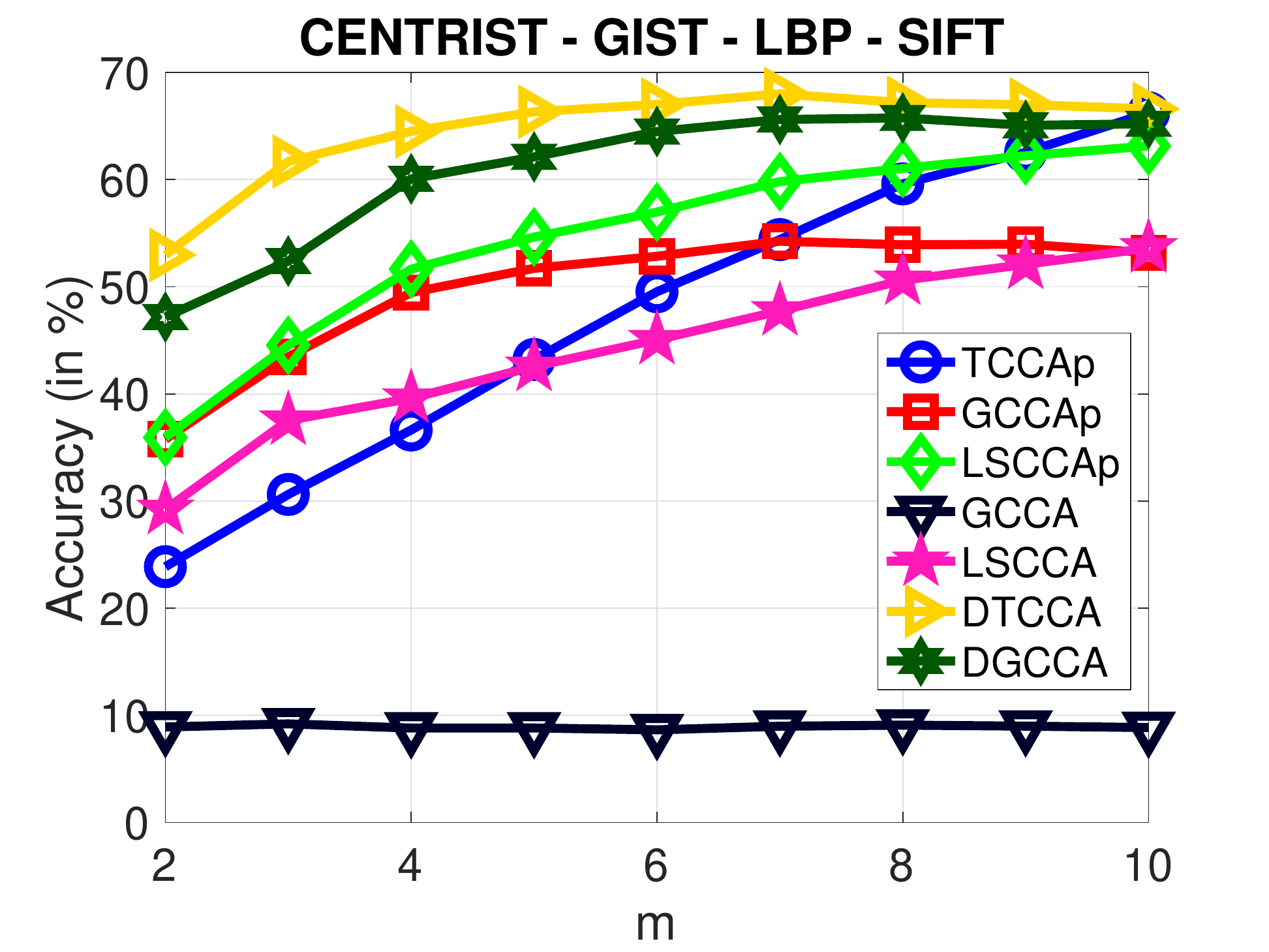} \\
	\includegraphics[width=0.45\textwidth]{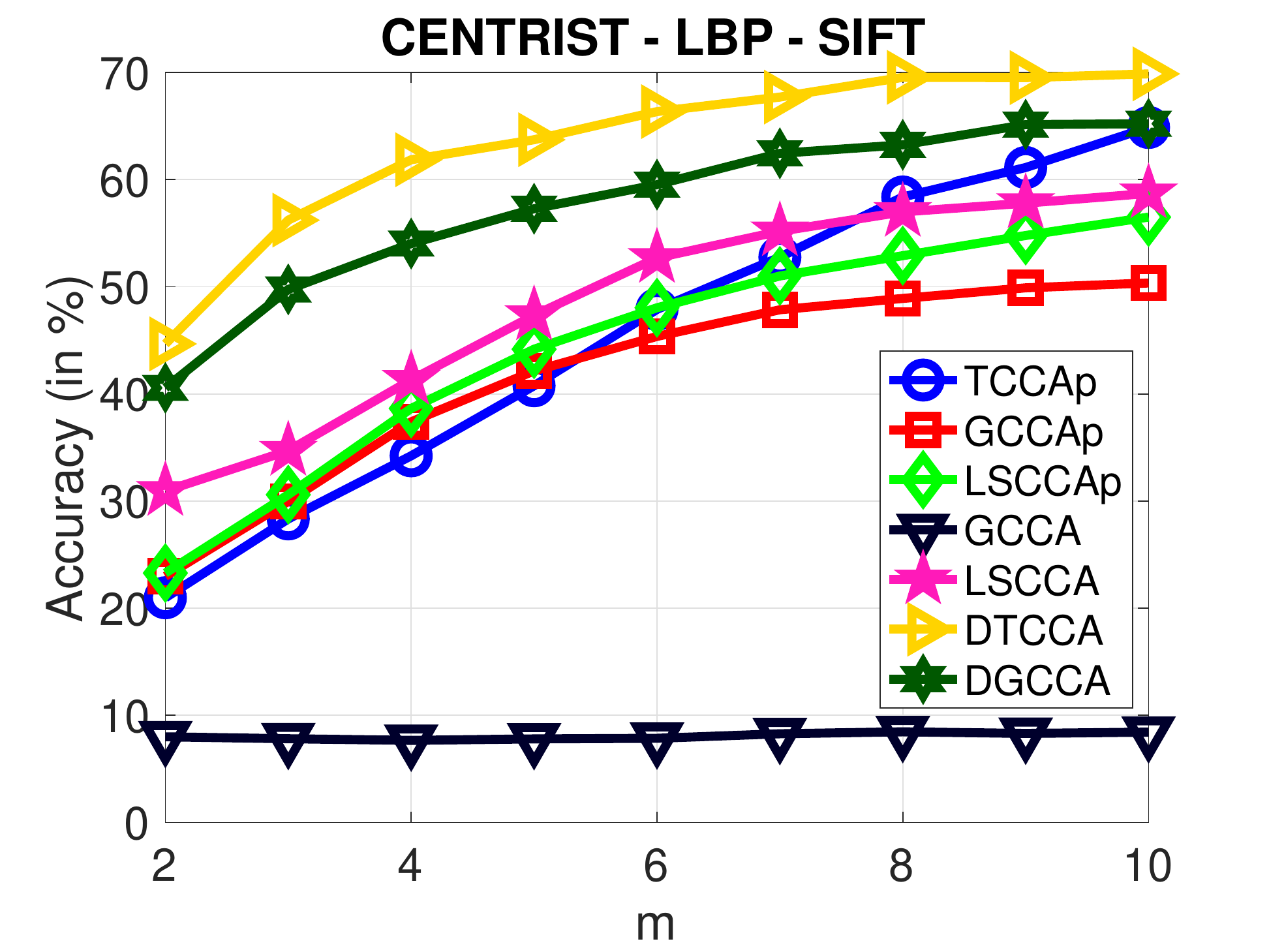} &
	\includegraphics[width=0.45\textwidth]{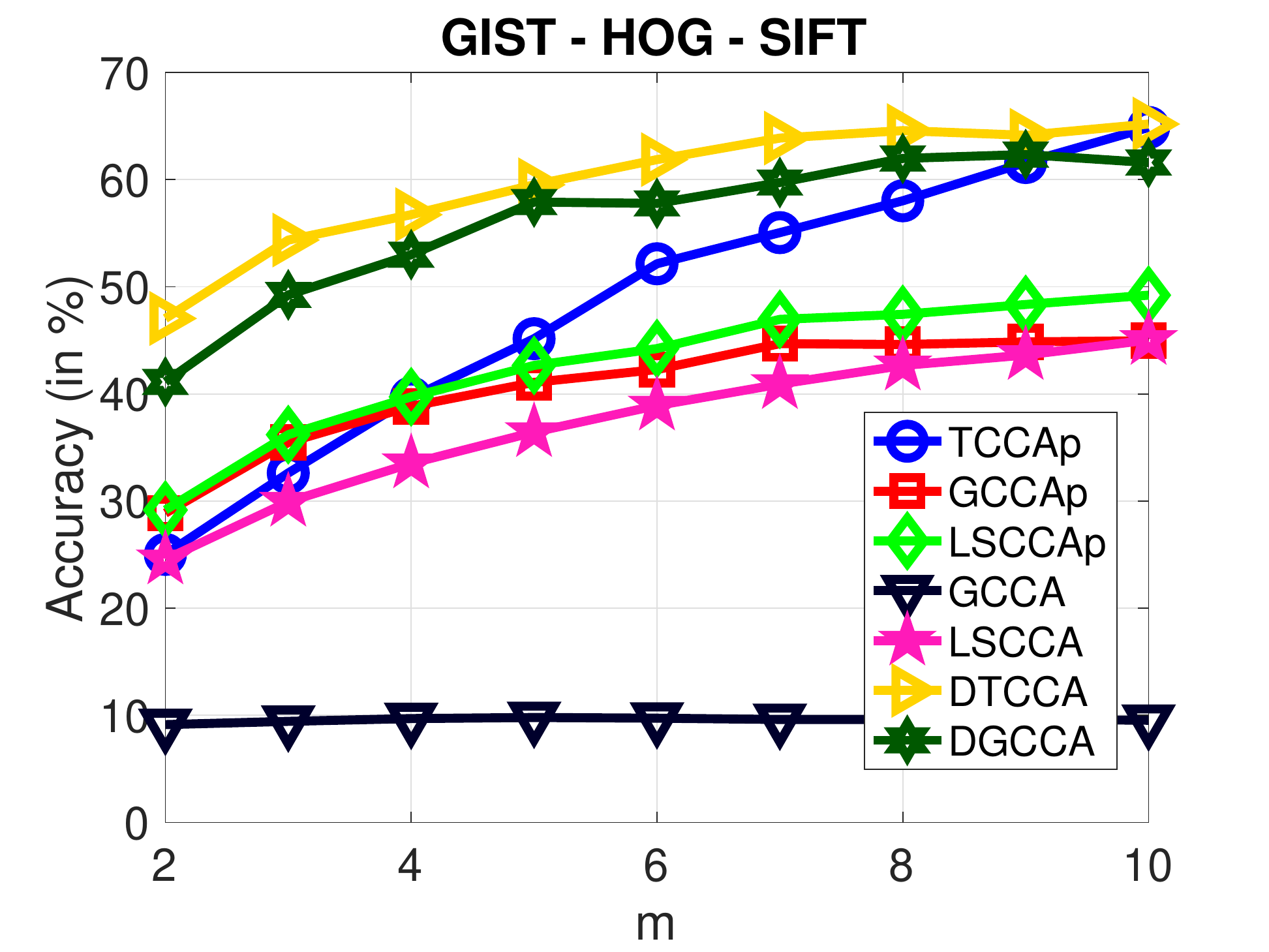} 
\end{tabular}
\caption{The accuracy of seven compared methods on Scene15 data by varying the dimension of the reduced space $m \in \{2,3,\ldots,10\}$.} \label{fig:scene15-dim}
\end{figure}

The 15 class scene data set was gradually built. The initial 8 classes were collected by Oliva and Torralba \cite{oliva2001modeling}, and then 5 categories were added by Fei-Fei and Perona \cite{fei2005bayesian}; finally, 2 additional categories were introduced by Lazebnik et al. \cite{lazebnik2006beyond}. The 15 scene categories are office, kitchen, living room, bedroom, store, industrial, tall building, inside cite, street, highway, coast, open country, mountain, forest, and suburb. Images in the data set are about $250\times 300$ resolution, with 210 to 410 images per class. This data set contains a wide range of outdoor and indoor scene environments. $4310$ images are used in this experiment. Five descriptors are used to generate the features of views including 254-d CENTRIST \cite{wu2008place}, 512-d GIST \cite{oliva2001modeling},
531-d LBP \cite{ojala2002multiresolution}, 360-d histogram of oriented gradient (HOG),
and 1000-d SIFT \cite{lazebnik2006beyond}. The same experiments are also conducted for Scene15 data. The classification accuracy in terms of the varied views, the dimension of common space, and view combinations are shown in Table \ref{tab:scene15-view}, Fig. \ref{fig:scene15-dim} 
and Table \ref{tab:scene15-all}, respectively. The similar observations as above two data sets can also be obtained on this data.

\begin{table}
\caption{The accuracy of $7$ compared methods over Scene15 data set by varying the number of views.} \label{tab:scene15-view}
\centering
\begin{tabular}{lccc}
	\hline
	& 3 views&4 views&5 views\\
	\hline
	GCCA& 8.69 $\pm$ 0.60& 9.02 $\pm$ 0.32& 9.44\\
	LSCCA& 47.60 $\pm$ 6.95& 44.72 $\pm$ 6.44& 41.07\\
	TCCA$_p$& 62.36 $\pm$ 4.91& 62.67 $\pm$ 4.24& 64.45\\
	GCCA$_p$& 51.27 $\pm$ 4.92& 51.74 $\pm$ 4.08& 51.27\\
	LSCCA$_p$& 55.52 $\pm$ 4.91& 59.08 $\pm$ 4.51& 62.72\\
	DGCCA& 63.96 $\pm$ 2.87& 64.48 $\pm$ 1.66& 64.71\\
	DTCCA& \textbf{65.80 $\pm$ 3.54}& \textbf{65.77 $\pm$ 1.99}& \textbf{67.33}\\
	\hline
\end{tabular}
\end{table}

\section{Conclusion} \label{sec:conclusions}
We propose DTCCA for dealing with multi-view extensions of TCCA by capturing the high-order statistics among all feature views and simultaneously learning the nonlinear transformations of each view in a unified model. Based on the experiments on various multi-view data sets, we have shown that DTCCA can obtain significant improvement comparing with TCCA and better or competitive results comparing with others with respect to the classification performance on test data. In addition, DTCCA does not need to apply preprocessing step to avoid the high computational complexity of TCCA on the high-dimensional input data in the case that the dimensionality of the latent subspace is relatively small. We observed that DTCCA can achieve consistently better results especially when the amount of training data is small. As a result, the combination of nonlinear transformation and maximizing high-order canonical correlations are important to improve the learning performance of multi-view data sets.

%
%


\end{document}